\documentclass[sn-mathphys]{sn-jnl}

\usepackage{booktabs}       
\usepackage{amsfonts}       
\usepackage{nicefrac}       
\usepackage{microtype}      
\usepackage{graphicx}
\usepackage{url}            

\usepackage{multirow}
\usepackage{subcaption}
\usepackage{xcolor}

\usepackage{natbib}
\setcitestyle{square,numbers}

\usepackage{lineno}

\jyear{2022}%

\theoremstyle{thmstyleone}%
\theoremstyle{thmstyletwo}%

\theoremstyle{thmstylethree}%

\begin{document}


\title[Neural network relief: a pruning algorithm based on neural activity]{Neural network relief: a pruning algorithm based on neural activity}


\author[1]{\fnm{Aleksandr} \sur{Dekhovich}}

\author[2]{\fnm{David} M.J. \sur{Tax}}

\author[1]{\fnm{Marcel} H.F \sur{Sluiter}}

\author*[3]{\fnm{Miguel} A. \sur{Bessa}}\email{miguel\_bessa@brown.edu}

\affil[1]{\orgdiv{Department of Materials Science and Engineering}, \orgname{Delft University of Technology}, \orgaddress{\street{Mekelweg 2}, \city{Delft}, \postcode{2628 CD},  \country{The Netherlands}}}

\affil[2]{\orgdiv{Pattern Recognition and Bioinformatics Laboratory}, \orgname{Delft University of Technology}, \orgaddress{\street{Van Mourik Broekmanweg 6}, \city{Delft}, \postcode{2628 XE}, \country{The Netherlands}}}

\affil*[3]{\orgdiv{School of Engineering}, \orgname{Brown University}, \orgaddress{\street{184 Hope St.}, \city{Providence}, \postcode{RI 02912}, \country{USA}}}


\abstract{Current deep neural networks (DNNs) are overparameterized and use most of their neuronal connections during inference for each task. The human brain, however, developed specialized regions for different tasks and performs inference with a small fraction of its neuronal connections. We propose an iterative pruning strategy introducing a simple importance-score metric that deactivates unimportant connections, tackling overparameterization in DNNs and modulating the firing patterns. The aim is to find the smallest number of connections that is still capable of solving a given task with comparable accuracy, i.e. a simpler subnetwork.  We achieve comparable performance for LeNet architectures on MNIST, and significantly higher parameter compression than state-of-the-art algorithms for VGG and ResNet architectures on CIFAR-10/100 and Tiny-ImageNet\footnote{Code implementation of the code is available at: \url{https://github.com/adekhovich/NNrelief}}. Our approach also performs well for the two different optimizers considered -- Adam and SGD. The algorithm is not designed to minimize FLOPs when considering current hardware and software implementations, although it performs reasonably when compared to the state of the art.}



\maketitle

\section{Introduction}\label{sec1}

Pruning is a common technique for neural network compression \cite{karnin1990simple}, where the main goal is to reduce memory and computational costs of inference. Pruning assumes particular relevance for deep neural networks because modern architectures involve several millions of parameters. Existing pruning methods are based on different strategies, e.g Hessian analysis \cite{lecun1990optimal, hassibi1993second}, magnitudes of weights  \cite{han2015learning}, data-driven approaches \cite{hu2016network,ancona2020shapley}, among others \cite{ullrich2017soft, molchanov2017variational}. Pruning can be done in one shot \cite{zhao2019variational} or in an iterative way \cite{frankle2018lottery}, and it is possible to prune connections \cite{lecun1990optimal, hassibi1993second, dong2017learning,han2015deep}, neurons \cite{molchanov2019importance, louizos2017bayesian, ancona2020shapley} or filters for convolutional layers \cite{he2017channel, he2018soft}. The typical pruning pipeline includes three stages: training the original network, pruning parameters and fine-tuning. Recently, interesting solutions for the third stage have been suggested that involve weight rewinding \cite{frankle2018lottery}  and learning rate rewinding \cite{renda2020comparing}. 

We have developed an algorithm that aims to prune a network without a significant decrease in accuracy after every iteration by keeping the signal in the network at some predefined level close to the original one. This contrasts with strategies that use a pre-defined ratio of parameters to prune \cite{hu2016network,ancona2020shapley,frankle2018lottery} which may lead to a drastic drop in accuracy for relatively high ratios. We look at the local behaviour of a particular connection and its contribution to the neuron, but not at the output or the loss function. Our aim is to deactivate unimportant connections for a given problem in order to free them for other tasks -- a crucial step towards novel architectural continual learning strategies \cite{mallya2018packnet, mallya2018piggyback}. Sparse architectures are also more robust to noisy data \cite{ahmad2019can} and exhibit benefits in the context of adversarial training \cite{ye2019adversarial, liao2022achieving}. Therefore, we focus on reducing the number of parameters of deep learning models targeting these scenarios, although we expect a reduction of the computational complexity (FLOPs) as well.


Our iterative pruning algorithm is based on an importance score metric proposed herein that quantifies the relevance of each connection to the local neuron behaviour. We show compressions of more than 50 times for VGG \cite{simonyan2014very} architectures on CIFAR-10 and Tiny-ImageNet datasets with a marginal drop of accuracy. We also apply our method to ResNets \cite{he2016deep}, achieving better parameter compression than state-of-the-art algorithms with a comparable decrease of accuracy on CIFAR-10 dataset. In addition, we visualise the effects of our pruning strategy on the information propagation through the network, and we observe a significant homogenization of the importance of the pruned neurons. We associate this homogenization with the notion of neural network relief: using fewer neuronal connections and distributing importance among them.

\section{Related work}
One of the first works eliminating unimportant connections in relatively small networks proposed analyzing the Hessian of the loss function \cite{lecun1990optimal, hassibi1993second} without network retraining. This idea was further developed for convolutional networks \cite{dong2017learning,lebedev2016fast}. However, computing second-order derivatives by calculating the Hessian is costly. 

The magnitude-based approach \cite{han2015learning} is simple and fast because it only involves pruning the weights with the smallest magnitude. This assumes that parameters with small magnitudes do not contribute significantly to the resulting performance. For convolutional layers, the sum of kernels' elements in the filters is considered and filters with the smallest sum are pruned \cite{li2016pruning}. Iterative magnitude pruning is applied for finding winning tickets \cite{frankle2018lottery} -- the minimal subnetwork that can be trained at least as well as the original one with the same hyperparameters. 

Other algorithms use input data to reduce the number of parameters. In \cite{hu2016network} pruning depends on the ratio of zero activations using ReLU function. So, the neurons that do not fire frequently enough are eliminated. A greedy approach is used for ThiNet \cite{luo2017thinet}, where channels that do not affect the resulting sum in convolutional layers are removed by solving an optimization problem. A similar algorithm is used in \cite{he2017channel}, but the optimization problem is solved with LASSO regression for determining ineffective channels. The property of convexity and sparsity that ReLU produces provides analytical boundaries for Net-trim \cite{aghasi2017net} by solving a convex
optimization problem. The game-theoretic approach with Shapley values \cite{ancona2020shapley} demonstrates good performance in a low-data regime, i.e. one-shot pruning without retraining, where neurons within one layer are considered as players in a cooperative game. 

NISP \cite{yu2018nisp} finds the contribution of neurons to the last layer before classification, while iSparse framework \cite{garg2020isparse} trains sparse networks eliminating the connections that do not contribute to the output. A pruning strategy where the connections' contribution is based on computing the loss function derivatives was also suggested in \cite{lee2018snip}.

Additional methods of interest include Bayesian weight pruning \cite{molchanov2017variational} and Bayesian compression \cite{louizos2017bayesian} to prune neurons and aim at computational efficiency. SSS \cite{huang2018data} introduces a scaling factor and sparsity constraints on this factor to scale neurons' or blocks' outputs. Similarly, SNLI \cite{ye2018rethinking} uses ISTA\cite{beck2009fast} to update the scaling parameter in Batch Normalization of convolutional layers to obtain a sparse representation. HRank \cite{lin2020hrank} calculates the average rank of feature maps and prunes filters with the lowest ones. FSABP \cite{geng2022pruning} finds filters that extract similar information and prunes them. As a result, filters that provide diversity in feature maps are retained. In the same way, CHIP \cite{sui2021chip} measures correlations between feature maps truncating feature maps with the lowest independence scores.

\section{Our pruning approach}
\label{section:approach}

Our goal is to eliminate connections in a layer that, on average,  provide a weak contribution to the next layer. We believe that a low magnitude of a weight does not mean that information passing through the connection has to have a negligible contribution. The converse is also assumed to be true: if a high weight magnitude is multiplied by a weak signal from the neuron (or even zero) then the contribution of that product is relatively insignificant in comparison with other input signals in the neuron, as long as this holds for most data. Therefore, our strategy contrasts significantly with magnitude-based pruning \cite{han2015learning}. 

\subsection{Fully connected layers}

\begin{figure}[ht]
    \centering
    \includegraphics[width=\textwidth]{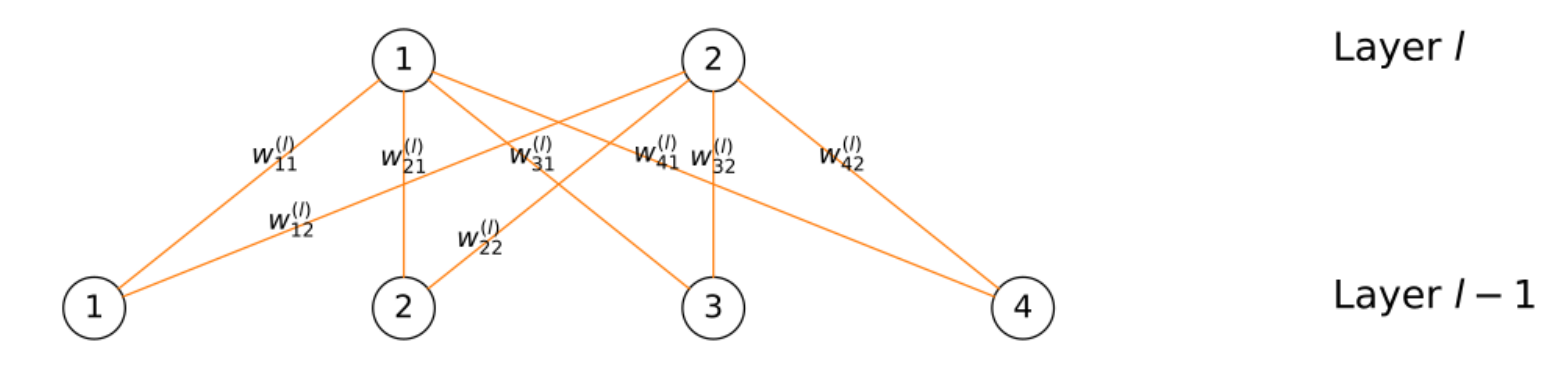}
    \caption{Neural network layers $l-1$ and $l$ with $m_{l-1} = 4$ and $m_l = 2$ neurons, respectively. The weights associated to a connection between neurons $i$ and $j$ in layer $l-1$ to $l$ are $w_{ij}^{(l)}$.}
    \label{fig:nn_schematic}
\end{figure}

Assume that we have a pruning set $\mathbf{X}^{(l-1)} = \{\mathbf{x}^{(l-1)}_1, \ldots, \mathbf{x}^{(l-1)}_N\}$ with $N$ samples, where each datapoint $\mathbf{x}^{(l-1)}_n = (x_{n1}^{(l-1)}, \ldots, x_{nm_{l-1}}^{(l-1)}) \in \mathbb{R}^{m_{l-1}}$ is the input for layer $l-1$ with dimension $m_{l-1}$ and where $\ 1 \le l \le L$ . We define the \textit{importance} of the connection between neuron $i$ of layer $l-1$ and neuron $j$ of layer $l$ as: 

\begin{equation}
    \label{eq:fc}
    s_{ij}^{(l)} = \frac{\overline{ \Big\lvert w_{ij}^{(l)} x_{i}^{(l-1)} \Big\rvert }}{\sum_{k=1}^{m_{l-1}} \overline{ \Big\lvert w^{(l)}_{kj} x_{k}^{(l-1)} \Big\rvert} + \Big\lvert b_j^{(l)}\Big\rvert },
\end{equation}

where $\overline{ \Big\lvert w_{ij}^{(l)}x_{i}^{(l-1)} \Big\rvert} = \frac{1}{N}\sum_{n=1}^N \Big\lvert w_{ij}^{(l)}x_{ni}^{(l-1)} \Big\rvert$, and $w_{ij}^{(l)}$ is the corresponding weight between neurons $i$ and $j$ (see Figure \ref{fig:nn_schematic}) and $b_j^{(l)}$ is the bias associated to neuron $j$. The importance score for the bias of neuron $j$ is $s^{(l)}_{m_{l-1}+1,j} = \frac{ \big\lvert b^{(l)}_j \big\rvert }{\sum_{k=1}^{m_{l-1}} \overline{ \big\lvert w^{(l)}_{kj} x_{k}^{(l-1)} \big\rvert}+ \big\lvert b_j^{(l)} \big\rvert }$. The denominator corresponds to the total importance in the neuron $j$ of layer $l$ that we denote as $S^{(l)}_j = \sum_{k=1}^{m_{l-1}} \overline{ \Big\lvert w^{(l)}_{kj} x_{k}^{(l-1)} \Big\rvert } + \Big\lvert b_j^{(l)} \Big\rvert, \ 1 \le j \le m_{l}$.

\begin{minipage}{\columnwidth}
    \begin{algorithm}[H]
      \begin{algorithmic}[1]
        \Function{FC pruning}{network, X, $\alpha$}
            \State{$\mathbf{X}^{(0)} \gets \mathbf{X}$}
            \For{every fc\_layer $l$ in FC\_Layers}
                \State{$\mathbf{X}^{(l)}$ $\gets$ fc\_layer$\big(\mathbf{X}^{(l-1)}\big)$}
                \For{ every neuron $j$ in fc\_layer $l$} 
                   \State{compute importance scores $s^{(l)}_{ij}$ for every incoming connection $i$ using \eqref{eq:fc}}.
                   \State{$\hat{s}^{(l)}_{ij} = Sort(s^{(l)}_{ij}, order=descending)$} 
                   \State{ $p_0 = \min\{p : \sum_{i=1}^p \hat{s}^{(l)}_{ij} \ge \alpha \}$}
                   \State{prune connections with importance score $s^{(l)}_{ij} < \hat{s}^{(l)}_{p_0j}$}
                \EndFor
            \EndFor
            \State{\Return pruned network}
        \EndFunction
      \end{algorithmic}
    \caption{Fully connected layers pruning}
    \label{alg:fc_pruning}
    \end{algorithm}
    \end{minipage}

\medskip

The idea behind our approach is to deactivate the connections $i = 1 \ldots, m_{l-1}$ that on average do not carry important information to the neuron $j$ of layer $l$ in comparison with other connections. As a result of Eq.\ref{alg:fc_pruning}, we simplify each neuron by using a smaller number of input connections, while causing minimal effective changes to the input signal of that neuron and subsequent minimal impact on the network.

\begin{figure}[ht]
  \begin{subfigure}[b]{0.5\linewidth}
    \includegraphics[width=\linewidth, height=2cm]{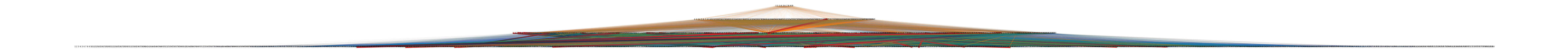}
    \caption{\small LeNet-300-100 pretrained on MNIST} 
  \end{subfigure} 
  \hfill
  \begin{subfigure}[b]{0.5\linewidth}
    \includegraphics[width=\linewidth, height=2cm]{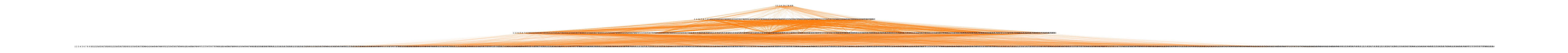}
    \caption{LeNet-300-100 pruned on MNIST} 
  \end{subfigure}
   \caption{LeNet-300-100 architecture on MNIST before and after pruning, where connections are coloured with respect to importance score: blue (least important) $\to$ red (most important).}
  \label{fig:lenet} 
\end{figure}

Figure \ref{fig:lenet} shows the importance score of every connection in the LeNet-300-100 \cite{lecun1998gradient} architecture before and after pruning when applied to the MNIST dataset. Connections are represented by thin blue lines when they have the lowest importance score $s_{ij}^{(l)}$, and go up to red thick lines when they are the most important. The figure demonstrates an interesting phenomenon besides the fact that there are much fewer connections after applying our pruning strategy: the importance scores of the connections become much closer to each other and there are no longer highly important and highly unimportant connections. We eliminate connections (contributors to the neuron's signal) that on average do not contribute in terms of the strength of the signal that they bring to the neuron for a given dataset. 

\subsection{Convolutional layers}

Our pruning approach for convolutional layers is similar to the one conducted on fully connected layers. We consider kernels and a bias in a particular filter as contributors to the signal produced by this filter.  

\begin{minipage}{\textwidth}
    \centering
    \begin{algorithm}[H]
      \begin{algorithmic}[2]
        \Function{CONV pruning}{network, X, $\alpha$}
        \State{$\mathbf{X}^{(0)} \gets \mathbf{X}$}
        \For{conv layer $l$ in CONV\_Layers}
            \State{$\mathbf{X}^{(l)} \gets$ conv\_layer$\big(\mathbf{X}^{(l-1)}\big)$}
            \For{ every filter $\mathbf{F}_j$ in conv\_layer $l$}
                \State{compute importance scores $s_{ij}$ $\forall$ kernel  $\mathbf{K}^{(l)}_{ij}$ and bias $b^{(l)}_j$ in filter $\mathbf{F}^{(l)}_j$ using \eqref{eq:conv}, \eqref{eq:bias}.}
                \State{$\hat{s}^{(l)}_{ij} = Sort(s^{(l)}_{ij}, order=descending)$} 
                \State{ $p_0 = \min\{p : \sum_{i=1}^p \hat{s}^{(l)}_{ij} \ge \alpha \}$}
                \State{prune kernel $\mathbf{K}^{(l)}_{ij}$ with importance score $s^{(l)}_{ij} < \hat{s}^{(l)}_{p_0j}$}    
            \EndFor  
        \EndFor
        \State{\Return pruned network}
        \EndFunction
      \end{algorithmic}
    \caption{Convolutional layers pruning}
    \end{algorithm}
\end{minipage}

Assume we have $m_{l-1}$-channelled input samples $\mathbf{X}^{(l-1)} = \{\mathbf{x}^{(l-1)}_1, \ldots, \mathbf{x}^{(l-1)}_N\}$, where  $\mathbf{x}^{(l-1)}_k = (x^{(l-1)}_{k1}, \ldots, x^{(l-1)}_{km_{l-1}}) \in \mathbb{R}^{m_{l-1} \times h_{l-1}^1 \times h^2_{l-1}}$, where $h^1_{l-1}$ and $h^2_{l-1}$ are the height and width of input images (or feature maps) for convolutional layer $l$. For every kernel $\mathbf{K}^{(l)}_{1j}, \mathbf{K}^{(l)}_{2j}, \ldots, \mathbf{K}^{(l)}_{m_l j}, \ \mathbf{K}^{(l)}_{ij} = (k^{(l)}_{ijqt}) \in \mathbb{R}^{r_l \times r_l}$, $1 \le q,t \le r_l$, $r_l$ is a kernel size, and a bias $b^{(l)}_j$ in filter $\mathbf{F}^{(l)}_j$, we define $\mathbf{\hat{K}}^{(l)}_{ij} = \Big( \Big\lvert k^{(l)}_{ijqt} \Big\rvert \Big)$ as a matrix consisting of the absolute values of the matrix $\mathbf{K}^{(l)}_{ij}$. 

Then we compute importance scores $s^{(l)}_{ij}, i \in \{1, 2, \ldots, m_l\}$ of kernels $\mathbf{K}^{(l)}_{ij}$ as follows:

\begin{align}
    \label{eq:conv}
    s^{(l)}_{ij} &= \frac{\frac{1}{N}\sum_{n=1}^N \Big\lvert\Big\lvert \mathbf{\hat{K}}^{(l)}_{ij} * \Big\lvert x^{(l-1)}_{ni} \Big\rvert \ \Big\rvert\Big\rvert_F}{S^{(l)}_{j}},\\
    \label{eq:bias}
    s^{(l)}_{m_l+1,j} &= \frac{ \Big\lvert b^{(l)}_j \Big\lvert\sqrt{h^{1}_{l} h^{2}_{l}}}{S^{(l)}_{j}}.
\end{align}
where $S^{(l)}_{j} = \sum_{i=1}^{m_{l-1}}\Big(\frac{1}{N}\sum_{n=1}^N \Big\lvert\Big\lvert \mathbf{\hat{K}}^{(l)}_{ij}* \Big\lvert x^{(l-1)}_{ni} \Big\rvert \ \Big\rvert\Big\rvert_F\Big) +  \Big\lvert b^{(l)}_j \Big\rvert \sqrt{h^{1}_{l} h^{2}_{l}}$ is the total importance score in filter $\mathbf{F}^{(l)}_j$ of layer $l$, and where $*$ indicates a convolution operation, and $\lvert \lvert\cdot \rvert \rvert_F$ the Frobenius norm. 

In Eq. \ref{eq:conv}, we compute the amount of information that every kernel produces on average, analogously to what we do in fully connected layers.

\section{Experiments}
\label{sec:experiments}
We test our pruning method for LeNet-300-100 and LeNet-5 \cite{lecun1998gradient} on MNIST, VGG-19 \cite{simonyan2014very} and VGG-like \cite{zagoruyko201592} on CIFAR-10/100 \cite{krizhevsky2009learning} and Tiny-ImageNet datasets. 
We evaluate our method on classification error, the percentage of pruned parameters \Big($\frac{\lvert w=0 \rvert}{\lvert w \rvert} (\%)$\Big) or the percentage of remaining parameters \Big($\frac{\lvert w \ne 0 \rvert}{\lvert w \rvert} (\%)$\Big) or compression rate \Big($\frac{\lvert w \rvert}{\lvert w \ne 0 \rvert} (\%)$\Big) and pruned FLOPs (floating-point operations), where $\lvert w \ne 0 \rvert$ refers to the number of unpruned connections. For the details about training and pruning hyperparameters, and FLOPs computation see Section \ref{sec:prunig_setups} and Section \ref{sec:flops}. We use the initialization method introduced in \cite{he2015delving} to initialize parameters. We run our experiments multiple times with different random initializations of the parameters. In our tables and figures, we present mean and standard deviation for the results averaged over these random initializations.

\subsection{LeNets}

We experiment with LeNets on the MNIST dataset. LeNet-300-100 is a fully connected network with 300 neurons and 100 neurons in two hidden layers respectively, and LeNet-5 Caffe, which is a modified version of \cite{lecun1998gradient}, has two convolutional layers followed by one hidden layer and an output layer. We perform iterative pruning by retraining the network starting from initial random parameters, instead of fine-tuning. Table \ref{tab:lenets_relu} shows that our approach is among the best for both LeNets in terms of pruned parameters.

\begin{table}[ht]
    \begin{center}
        \caption{Results for LeNet-300-100 and LeNet-5 trained and pruned on MNIST. For pruning, 1000 random training samples are chosen, $\alpha_{\text{fc}}=0.95, \ \alpha_{\text{conv}}=0.9$.}
        \label{tab:lenets_relu}
        \begin{tabular}{@{}llll@{}} 
        \toprule
        Network & Method & Error $(\%)$ & Parameters retained (\%)\\ 
        \midrule
        \multirow{4}{3cm}{LeNet-300-100} 
                          & DNS \cite{guo2016dynamic} & 1.99 & 1.79\\
                          & L-OBS \cite{dong2017learning} & 1.96 & 1.5\\
                          & SWS \cite{ullrich2017soft} & 1.94 & 4.3\\
                          & Sparse VD \cite{molchanov2017variational} & \textbf{1.92} & \textbf{1.47}\\
                          \cline{2-4}
                          &\textbf{NNrelief (ours)} & 1.98 $\pm$  0.07 & 1.51 $\pm$ 0.07\\
        \midrule
        \multirow{4}{3cm}{LeNet-5} 
                        & DNS \cite{guo2016dynamic}  & 0.91 & 0.93\\
                        &L-OBS \cite{dong2017learning}  & 1.66 & 0.9\\
                        &SWS \cite{ullrich2017soft} & 0.97 & 0.5\\
                        & Sparse VD \cite{molchanov2017variational}  & \textbf{0.75} & \textbf{0.36}\\
                        \cline{2-4}
                        &\textbf{NNrelief (ours)}  & 0.97 $\pm$ 0.05 & 0.65 $\pm$ 0.02\\
        \botrule
                             
        \end{tabular}
    \end{center}
\end{table}

\subsection{VGG}

We perform our experiments on VGG-13 and VGG-like (adapted version of VGG-16 for CIFAR-10 dataset that has one fully connected layer less) on CIFAR-10, CIFAR-100 and Tiny-ImageNet datasets. We also show in this section the effect of two optimizers -- Adam \cite{kingma2014adam} and SGD and two weight decay values on our pruning technique.

\begin{figure}[ht!]
    \begin{subfigure}{0.47\textwidth}
        \includegraphics[width=\textwidth]{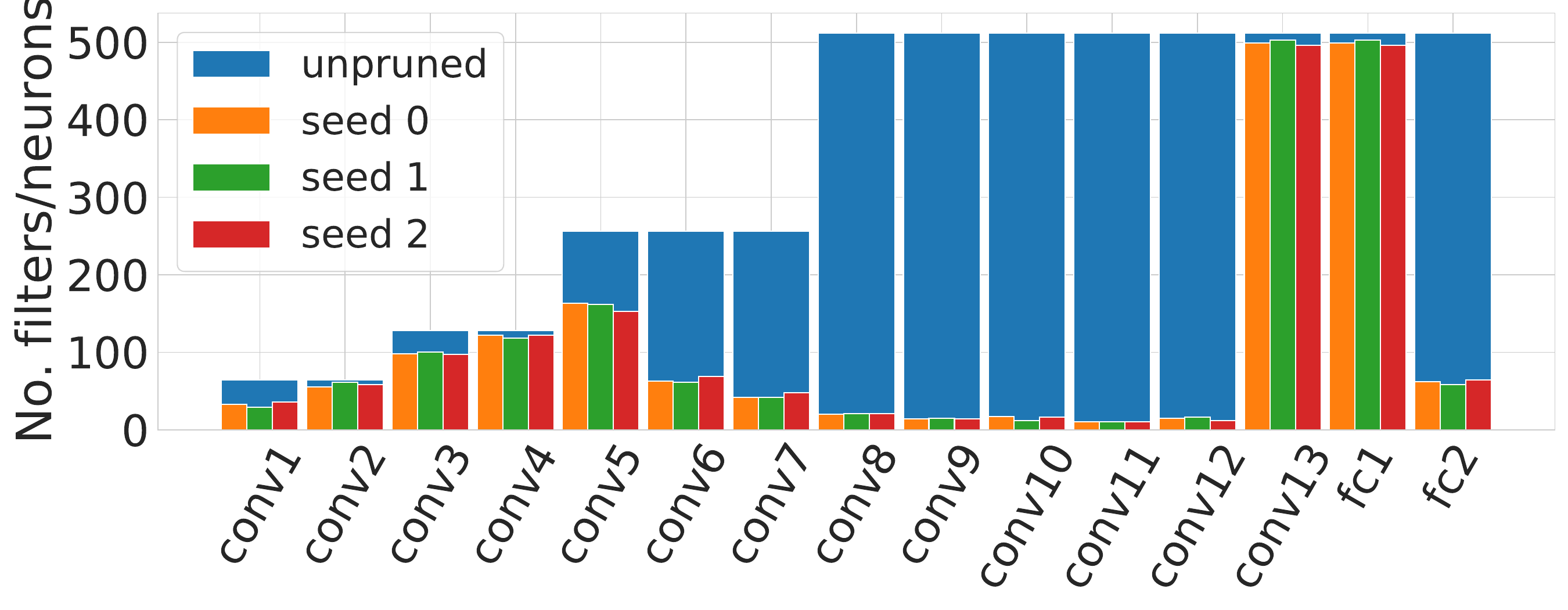}
        \caption{An architecture before and after pruning by seed.}
        \label{fig:vggnet_arch}
    \end{subfigure}
    \begin{subfigure}{0.47\textwidth}
        \includegraphics[width=\textwidth]{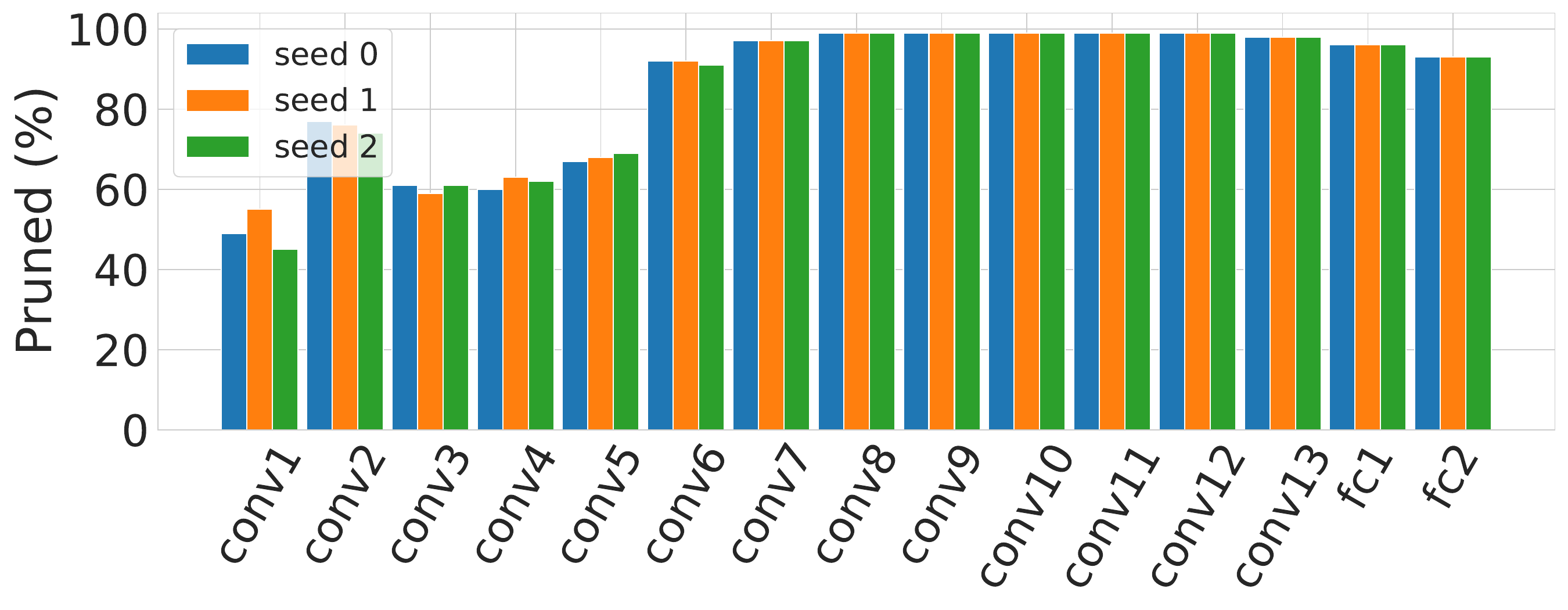}
        \caption{Sparsity by layer after pruning using different seeds.}
        \label{fig:vggnet_sparsity}
    \end{subfigure}
    \caption{Architecture structure for VGG-like on CIFAR-10 with Adam optimizer considering three random initializations.}
    \label{fig:vggnet_cifar10}
\end{figure}

\bmhead{CIFAR-10/100} Table \ref{tab:vgg_results} shows that after the reduction of the weights to less than $2\%$ of the original number, the accuracy only drops by $0.1\%$, and that our approach outperforms others both in terms of pruned parameters and pruned FLOPs. In Figure \ref{fig:vggnet_cifar10}, we show the final architecture after pruning and the level of sparsity by layer. It can be observed that NNrelief with Adam optimizer achieves a high level of filter sparsity even though it prunes kernels. 

\begin{table}[ht]
    \begin{center}
        \caption{Results for VGG-like trained on CIFAR-10 with Adam optimizer. During retraining 60 epochs are used; the learning rate is decreased by 10 every 20 epochs. For NNrelief (ours) $\alpha_{\text{conv}} = \alpha_{\text{fc}}=0.95$ and 1000 samples are chosen for the importance scores computation. Sign '-' means that accuracy has increased after pruning.}
        \label{tab:vgg_results}
        \begin{tabular}{ @{} p{0.6in} l p{0.5in} p{0.45in} p{0.65in} p{0.65in} @{}} 
        \toprule
        Network & Method & Acc (\%) (baseline) & Acc drop (\%) & Parameters retained (\%) & FLOPs pruned (\%) \\
        \midrule
        \multirow{7}{1cm}{VGG-like}
            & Pruning \cite{li2016pruning} & 93.25 & -0.15 & 36 & 34.2 \\
            & Sparse VD \cite{molchanov2017variational} & 92.3 & 0.0 & 2.1 & N/A \\
            & BC-GNJ \cite{louizos2017bayesian} & 91.6 & 0.2 & 6.7 & 55.6\\
            & BC-GHS \cite{louizos2017bayesian} & 91.6 & 0.6 & 5.5 & 61.7\\
            & SNIP \cite{lee2018snip} & 91.7 & \textbf{-0.3} & 3.0 & N/A\\
            & HRank \cite{lin2020hrank} & 93.96 & 1.62 & 17.9 & 65.3\\
            & CHIP \cite{sui2021chip} & 93.96 & 0.24 & 16.7 & 66.6\\
            \cline{2-6}
            &\textbf{NNrelief (ours)} & 92.5 & 0.1  & \bf 1.92 $\pm$ 0.02 & \textbf{75.5 $\pm$ 0.4}\\
        \botrule
        \end{tabular}
    \end{center}
\end{table}

\begin{figure}[ht!]
    \centering
    \includegraphics[width=\columnwidth]{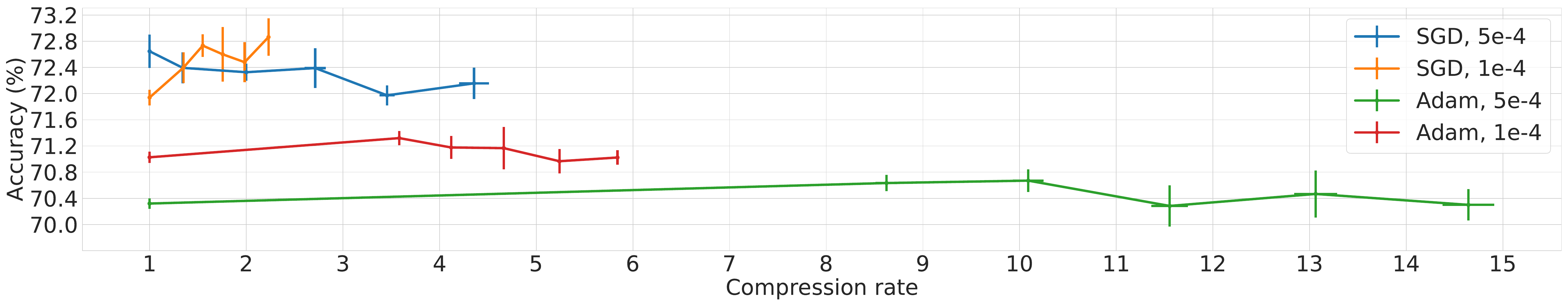}
     \caption{Results for VGG-13 over three seeds; mean values are used to compute dots and standard deviation are shown with error bars. We compare two optimizers, SGD and Adam, and two different values of weight decay for evaluation after 5 pruning iterations.}
    \label{fig:vgg13_results}
\end{figure}

\begin{figure}[ht!]
  \begin{subfigure}[b]{0.5\textwidth}
    \includegraphics[width=\textwidth]{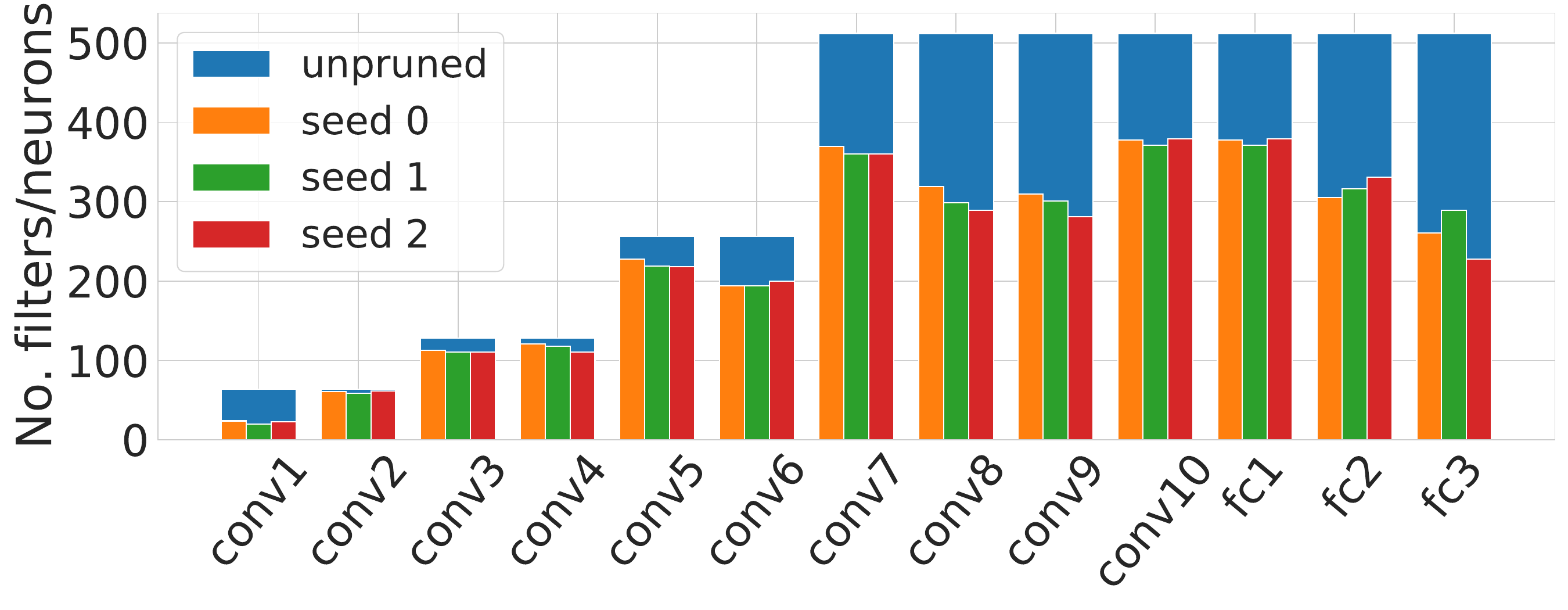}
    \caption{SGD, weight decay = $5 \cdot 10^{-4}$}
  \end{subfigure} 
  \begin{subfigure}[b]{0.5\textwidth}
    \includegraphics[width=\textwidth]{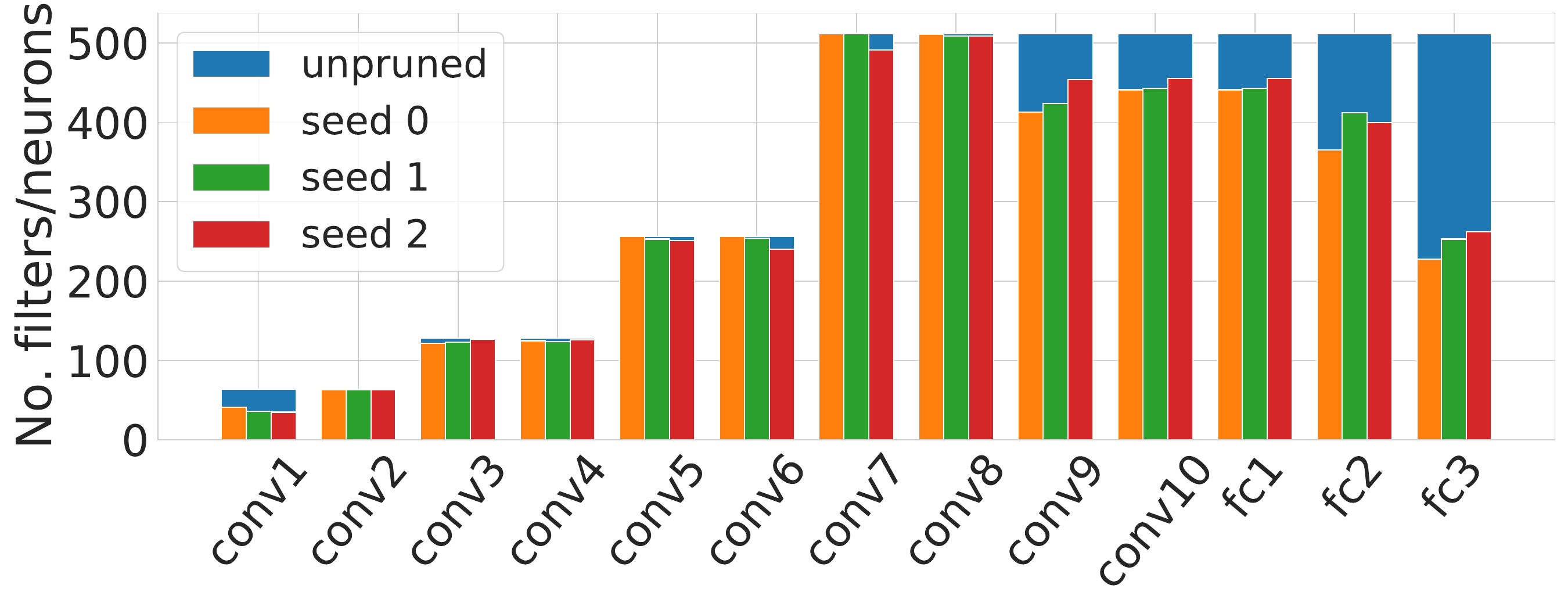}
    \caption{SGD, weight decay = $10^{-4}$}
  \end{subfigure} 
  \begin{subfigure}[b]{0.5\textwidth}
    \includegraphics[width=\textwidth]{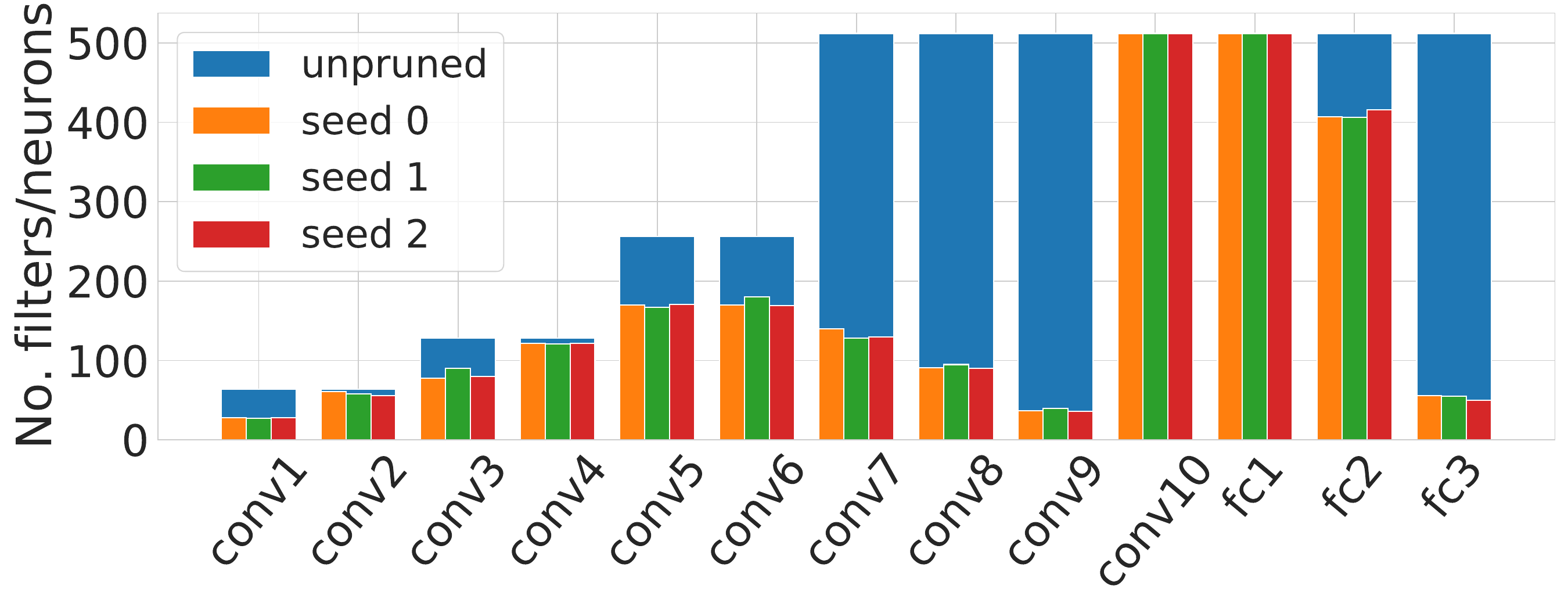}
    \caption{Adam, weight decay = $5 \cdot 10^{-4}$}
  \end{subfigure}
  \hfill
  \begin{subfigure}[b]{0.5\textwidth}
    \includegraphics[width=\textwidth]{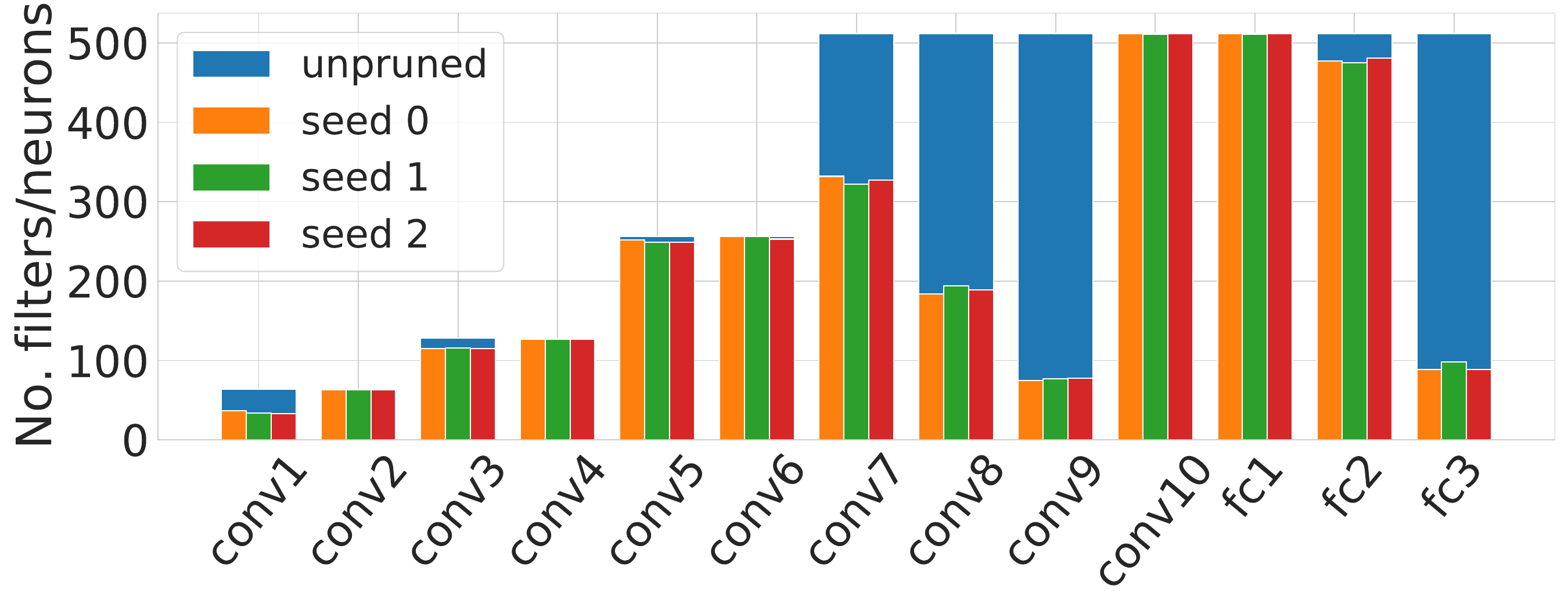}
    \caption{Adam, weight decay = $10^{-4}$}
  \end{subfigure}
  \caption{VGG-13 architecture on CIFAR-100 trained with SGD (top) and Adam (bottom), and weight decay equal to $5 \cdot 10^{-4}$ (left) and $10^{-4}$ (right) and pruned with 5 iterations. The results for three different seeds are presented.}
  \label{fig:vgg13_sgd_adam_weight-decay} 
\end{figure}

We also perform pruning for standard VGG-13 on CIFAR-100 using 5 iterations, and considered three randomization seeds to provide reasonable statistical significance. The results from the different optimizers are presented in Figure \ref{fig:vgg13_results}. The results show that Adam seems to perform the compression more aggressively, resulting in a higher compression rate, and also a slightly lower accuracy. The compression of about 14.5$\times$ after the fifth iteration without loss in accuracy is presented for Adam and weight decay $5\cdot 10^{-4}$, while SGD allows to train a network with higher accuracy, but after 5 iterations we obtain smaller compression (about $4.5\times$) and the decrease of accuracy by $0.5\%$ with the same weight decay. We obtain a lower compression rate for both optimizers when the weight decay is lower ($10^{-4}$). Figure \ref{fig:vgg13_sgd_adam_weight-decay} shows the final architecture and sparsity that we achieve with Adam optimizer at the end of the pruning procedure. We observe a high level of filter sparsity with Adam, as also reported in \cite{mehta2019implicit}, but good compression is also obtained for the SGD optimizer.

\bmhead{Tiny-ImageNet} We also apply our approach to Tiny-ImageNet dataset that is a subset of the ImageNet dataset with 200 classes and an image spatial resolution of 64$\times$64. Following SNIP strategy, we use strides $[2, 2]$ in the first convolutional layer to reduce the size of images. The results are reported in Table \ref{tab:vgg-like_tiny-imagenet}. We observe a compression by more than 40 times almost without any loss of accuracy ($-0.03\%$) using Adam optimizer.

\begin{table}[ht!]
    \begin{center}
        \caption{Results for VGG-like trained on Tiny-ImageNet with Adam optimizer. During retraining 50 epochs are used; the learning rate is decreased by 10 every 20 epochs. For NNrelief (ours) $\alpha_{\text{conv}} = \alpha_{\text{fc}}=0.95$ and 2000 samples are chosen for the importance scores computation.}
        \begin{tabular}{ @{} l l p{0.5in} p{0.45in} p{0.65in} p{0.65in} @{} } 
        \toprule
        Network & Method & Acc (\%) (baseline) & Acc drop (\%) & Parameters retained (\%) & FLOPs pruned (\%) \\ 
        \midrule
        \multirow{2}{1.2cm}{VGG-like}
            & SNIP \cite{lee2018snip} & 45.14 & 0.87 & 5.0 & N/A\\
            \cline{2-6}
            &\textbf{NNrelief (ours)} & 45.63 & 0.03 & \textbf{2.32 $\pm$ 0.06} & \textbf{75 $\pm$ 0.32}\\
        \botrule
        \end{tabular}
        \label{tab:vgg-like_tiny-imagenet}
    \end{center}
\end{table}

\subsection{ResNets}

\begin{figure}[ht!]
    \begin{subfigure}{0.5\textwidth}
        \includegraphics[width=\textwidth]{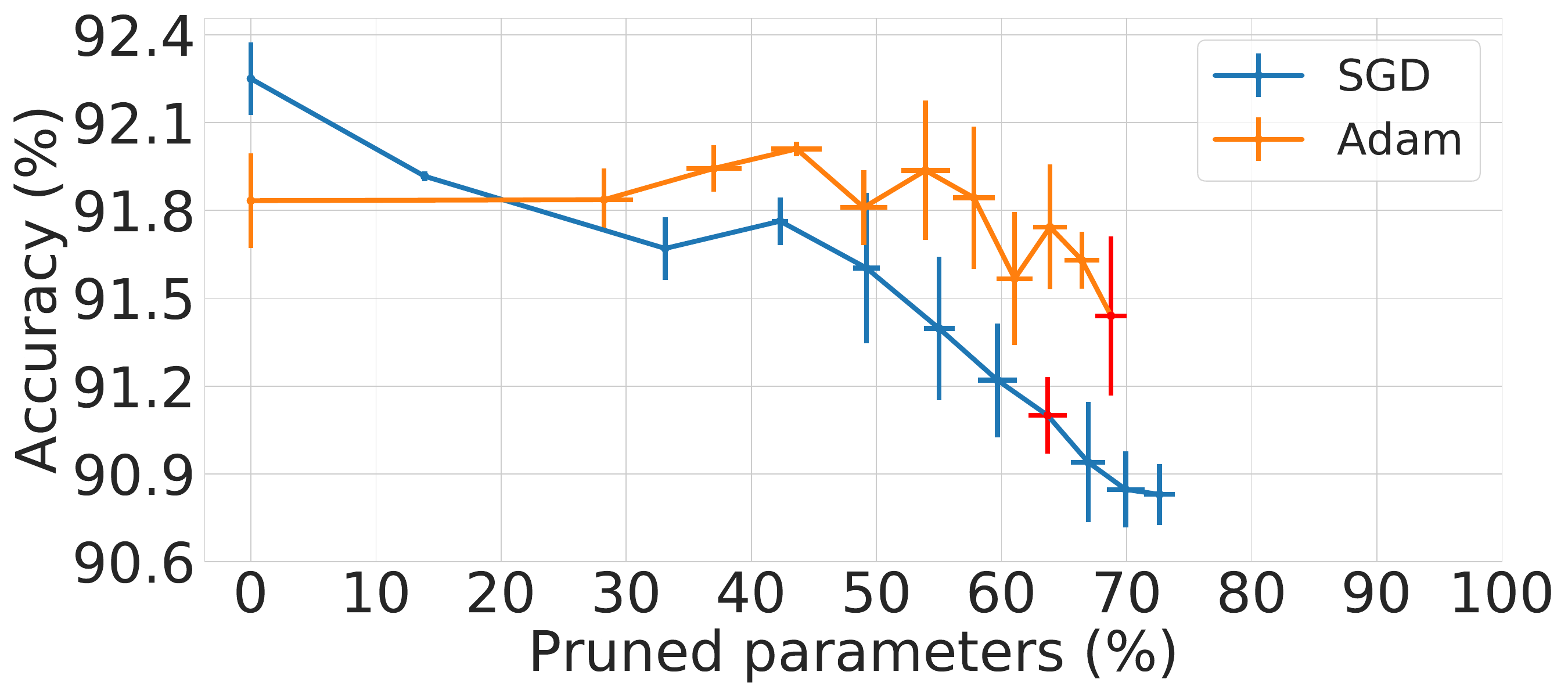}
        \caption{CIFAR-10}
        \label{fig:resnet20_cifar10}
    \end{subfigure}
    \begin{subfigure}{0.5\textwidth}
        \includegraphics[width=\textwidth]{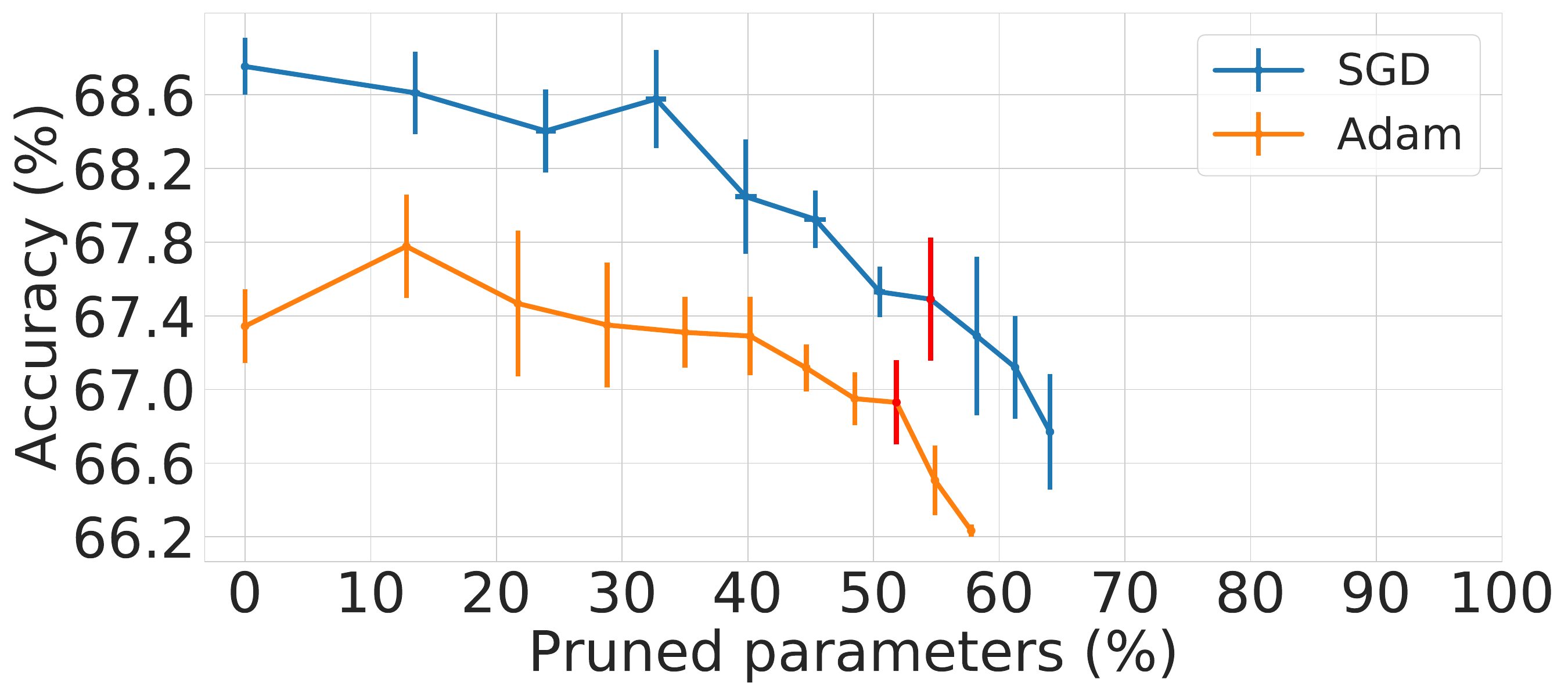}
        \caption{CIFAR-100}
        \label{fig:resnet20_cifar100}
    \end{subfigure}
    \begin{subfigure}{0.5\textwidth}
        \includegraphics[width=\textwidth]{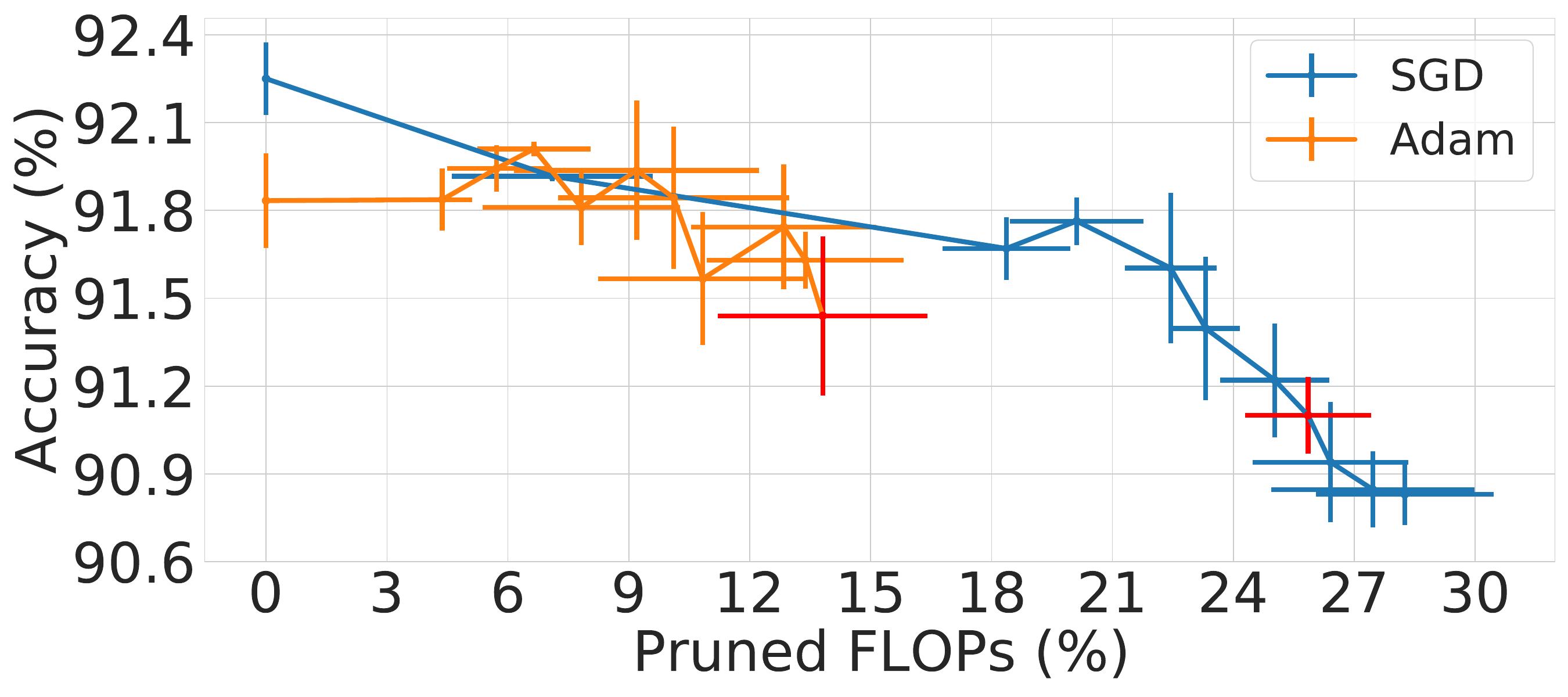}
        \caption{CIFAR-10}
        \label{fig:resnet20_cifar10_flops}
    \end{subfigure}
    \begin{subfigure}{0.5\textwidth}
        \includegraphics[width=\textwidth]{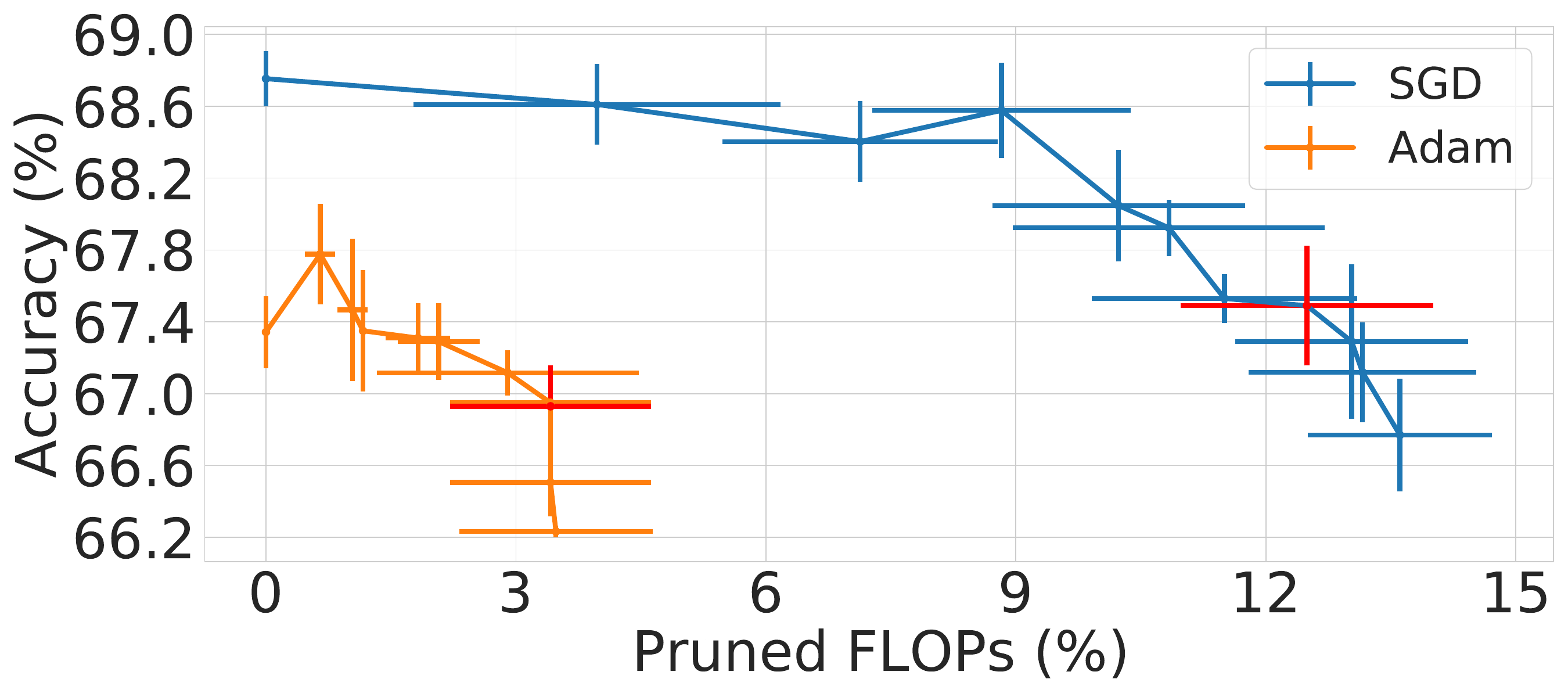}
        \caption{CIFAR-100}
        \label{fig:resnet20_cifar100_flops}
    \end{subfigure}
    \caption{ResNet-20 on CIFAR-10 and CIFAR-100 by iteration averaged over three random initializations, the error bars represent standard deviation for these three runs. The red dot corresponds to the selected best iteration.}
    \label{fig:resnet20_cifar}
\end{figure}

In addition, we test our approach on ResNet architectures for CIFAR-10/100 and Tiny-ImageNet datasets. Our main objective is to prune as many parameters as we can without significant loss of accuracy. For ResNet-20/56 on CIFAR-10, we perform iterative pruning over 10 iterations, training the model with SGD and Adam. In Table \ref{tab:resnet_cifar10_results} we compare our results with other approaches, and Figure \ref{fig:resnet20_cifar} displays the pruning history. 
We achieve a higher percentage of pruned parameters for ResNet-20 and ResNet-56, using both optimizers with comparable final accuracy. 

\begin{table}[ht!]
    \begin{center}
        \caption{Results for ResNets trained on CIFAR-10 with SGD and Adam optimizers. During retraining 60 epochs are used; the learning rate is decreased by 10 every 20 epochs. For NNrelief (ours) $\alpha_{\text{conv}} = 0.95, \alpha_{\text{fc}}=0.99$.} \label{tab:resnet_cifar10_results}
        \begin{tabular}{ @{} p{0.25in} p{0.95in} p{0.65in} p{0.65in} p{0.7in} p{0.65in} @{} } 
        \toprule
        ResNet & Method & Acc (\%) (baseline) & Acc drop (\%)& Parameters pruned (\%) & FLOPs pruned (\%)\\ 
        \midrule
        \multirow{6}{0.1cm}{20} 
            & SFP \cite{he2018soft} & 92.2 & 1.37 & 30 & 42.2 \\
            & SNLI \cite{ye2018rethinking} & 92.0  & 1.1 & 37.2 & N/A \\
            & SSS \cite{huang2018data} & 92.8 & 2 & 45 & \textbf{60}\\
            & CNN-CFC \cite{li2019compressing} & 92.2 & 1.07 & 42.75  & 41.6\\
            \cline{2-6}
            &\textbf{NNrelief} (SGD) & 92.25 $\pm$ 0.12 & 1.15 $\pm$ 0.13 & \textbf{63.68 $\pm$ 1.52} &  25.85 $\pm$ 1.57\\
            &\textbf{NNrelief} (Adam) & 91.83 $\pm$ 0.16 & \textbf{0.39 $\pm$ 0.27} & \textbf{68.75 $\pm$ 1.26} &  13.81 $\pm$ 2.6\\
        \midrule
        \multirow{9}{0.1cm}{56} 
            & Pruning-B \cite{li2016pruning} & 93.04 & -0.02 & 13.7 & 27.6 \\        
            & SFP \cite{he2018soft} & 93.59 & -0.19 & 30 & 41.1 \\
            & NISP \cite{yu2018nisp} & N/A  & 0.03 & 42.6 & 43.61 \\
            & \multirow{2}{2cm}{CNN-CFC \cite{li2019compressing}}                
                  & 93.14 & \textbf{-0.24} & 43.09 & 42.78 \\
                  && 93.14 & 1.22 & 69.74 & 70.9\\
            & \multirow{2}{1.5cm}{HRank \cite{lin2020hrank}}                        
                  & 93.36 & 0.09 & 42.4 & 50 \\
                  && 93.36 & 2.54 & 68.1 & \textbf{74.1}\\
            & CHIP \cite{sui2021chip} & 93.26 & 1.21  & 71.8 & 72.3 \\
            \cline{2-6}
            &\textbf{NNrelief} (SGD) & 93.6 $\pm$ 0.08 & 1.14 $\pm$ 0.22 & \textbf{76.2 $\pm$ 0.44} &  35.8 $\pm$ 2.69 \\
            &\textbf{NNrelief} (Adam) & 92.8 $\pm$ 0.08 & 0.03 $\pm$ 0.23 & \textbf{75.95 $\pm$ 0.22} &  32.5 $\pm$ 0.31 \\
        \bottomrule
        \end{tabular}
    \end{center}
\end{table}

\begin{figure}[ht!]
    \centering
    \includegraphics[width=\columnwidth]{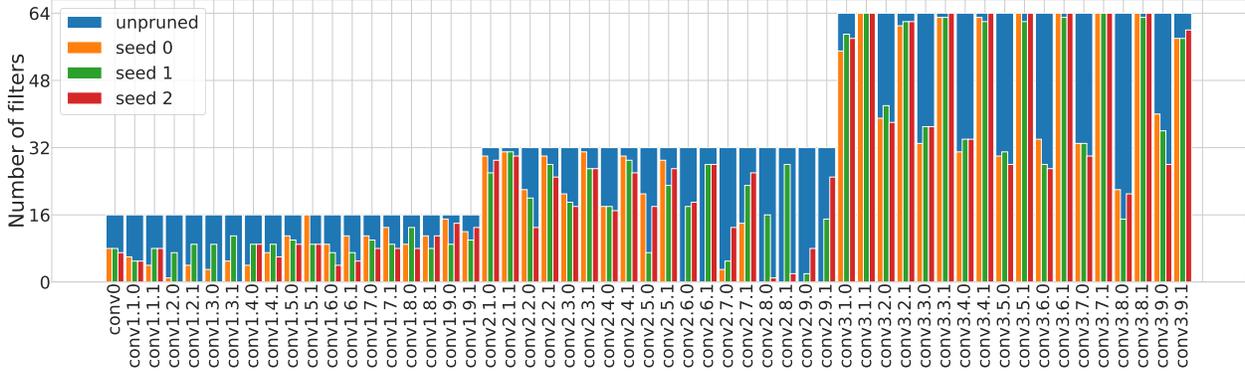}
    \caption{ResNet-56 architecture on CIFAR-10 by seed with SGD optimizer.}
    \label{fig:resnet56_cifar10_arch}
\end{figure}

A notable observation for some seeds when training with SGD is that separate convolutional blocks are pruned and only residual blocks remain, i.e. the signal propagates through skip connections and not through main convolutional layers since they do not contribute to the sum (Figure \ref{fig:resnet56_cifar10_arch}).

\begin{figure}[ht!]
    \begin{subfigure}{0.5\textwidth}
        \includegraphics[width=\textwidth]{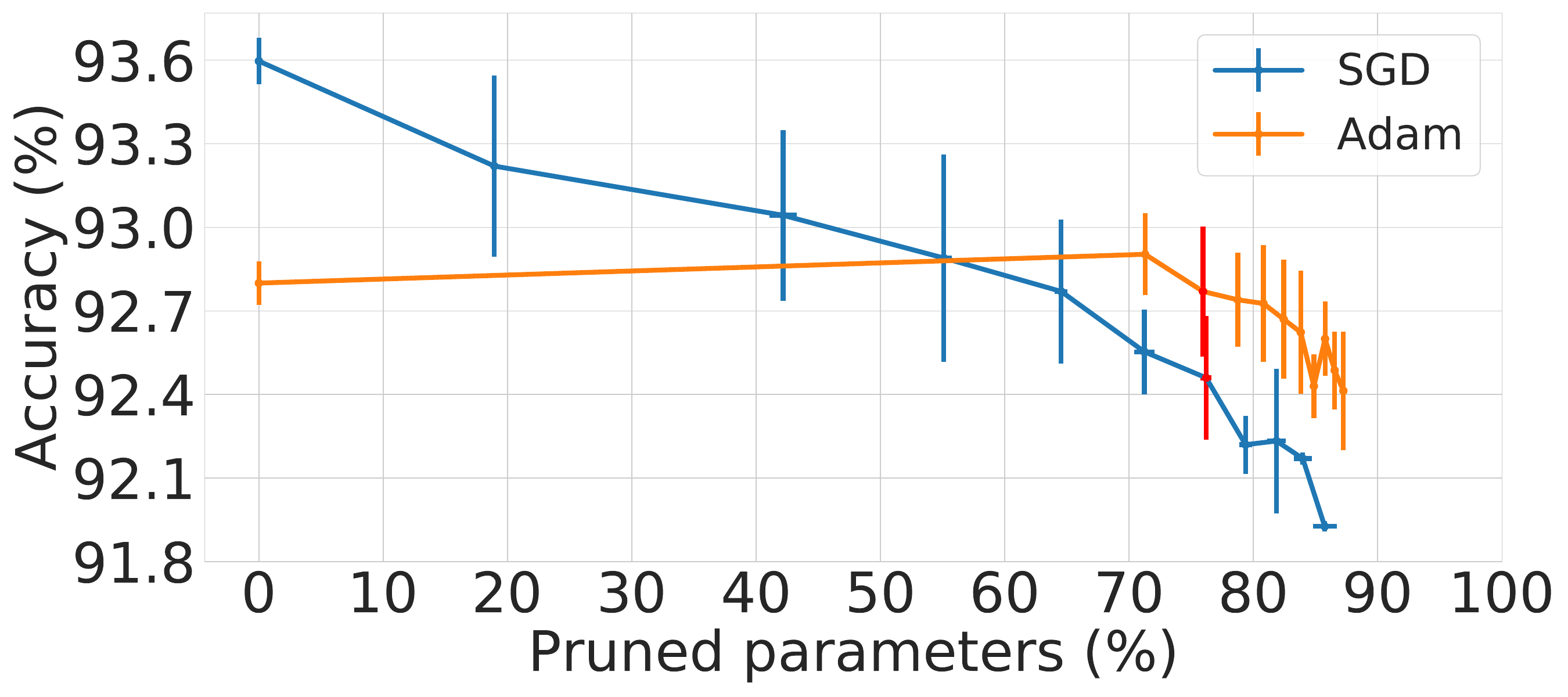}
        \caption{CIFAR-10}
        \label{fig:resnet56_cifar10}
    \end{subfigure}
    \begin{subfigure}{0.5\textwidth}
        \includegraphics[width=\textwidth]{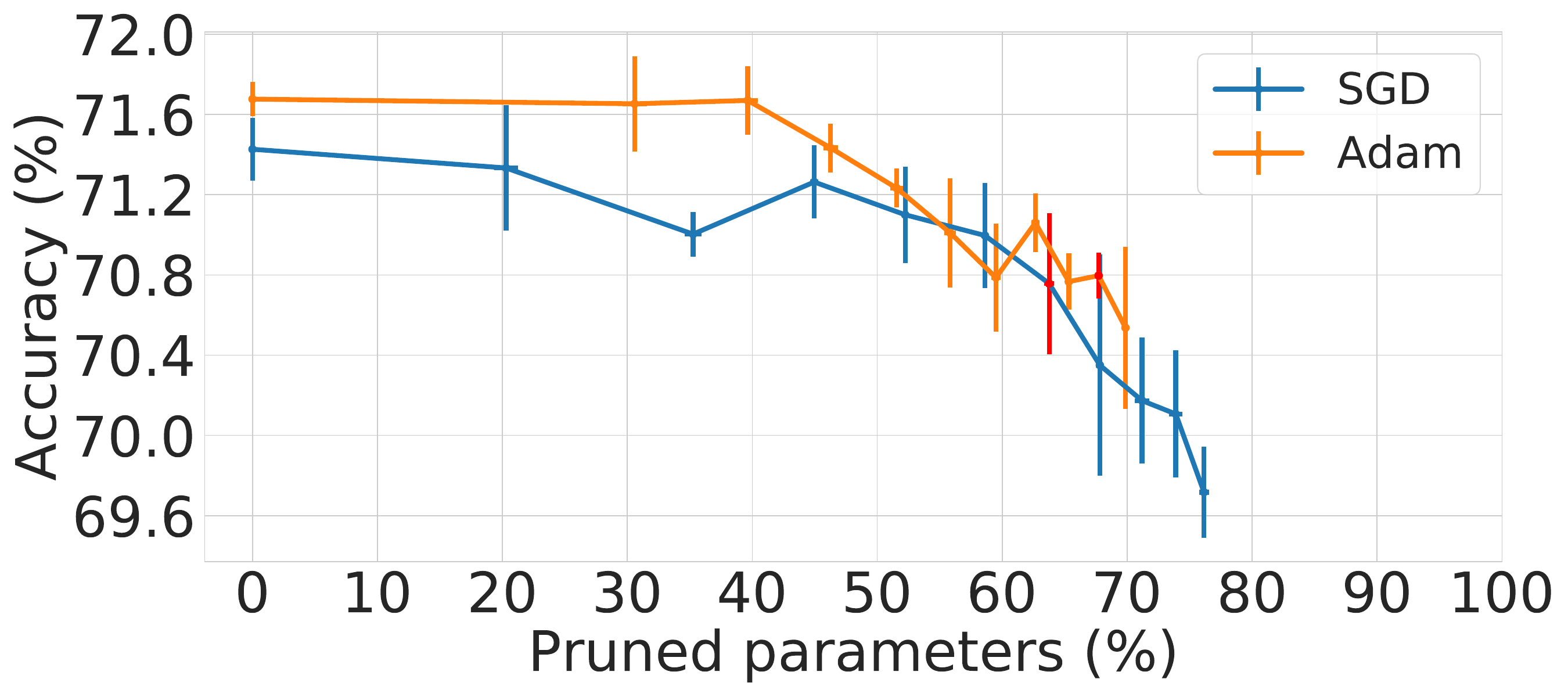}
        \caption{CIFAR-100}
        \label{fig:resnet56_cifar100}
    \end{subfigure}
    \begin{subfigure}{0.5\textwidth}
        \includegraphics[width=\textwidth]{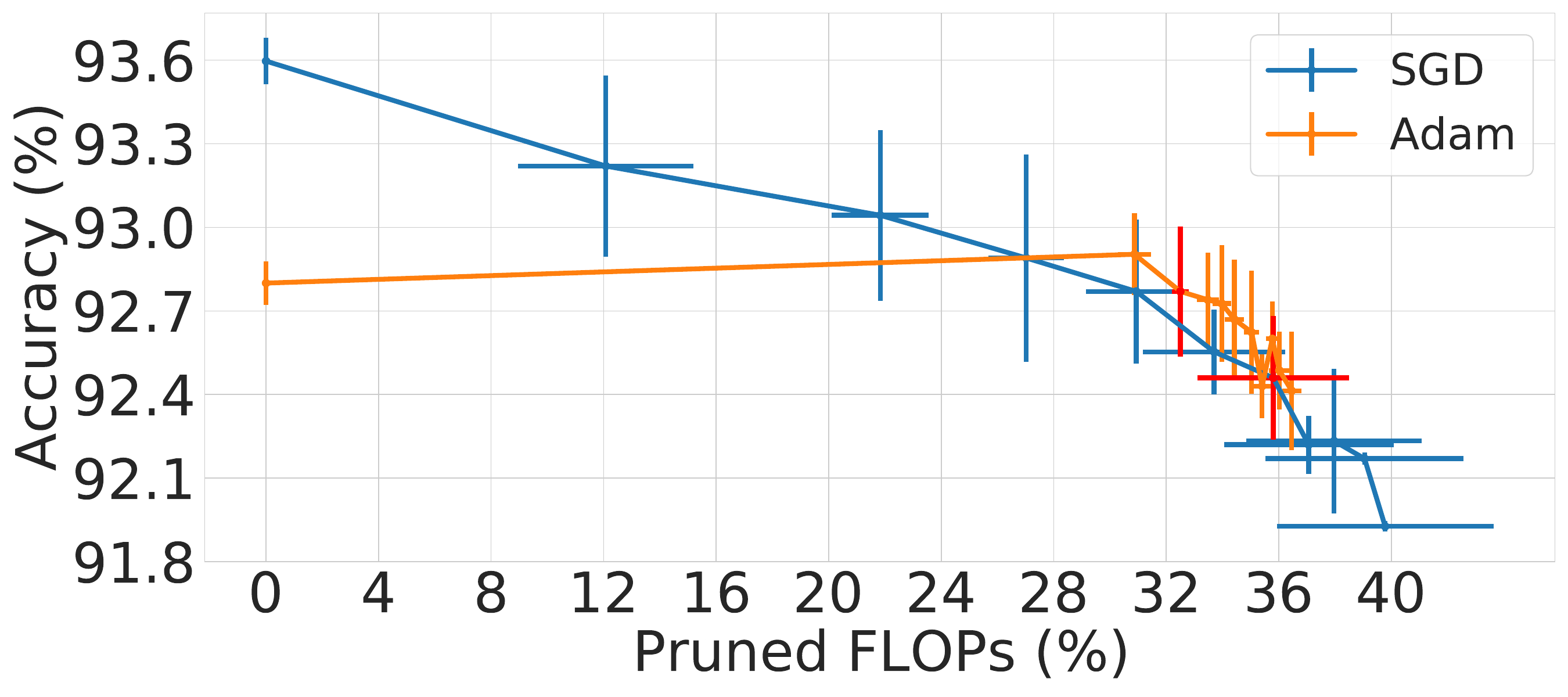}
        \caption{CIFAR-10}
        \label{fig:resnet56_cifar10_flops}
    \end{subfigure}
    \begin{subfigure}{0.5\textwidth}
        \includegraphics[width=\textwidth]{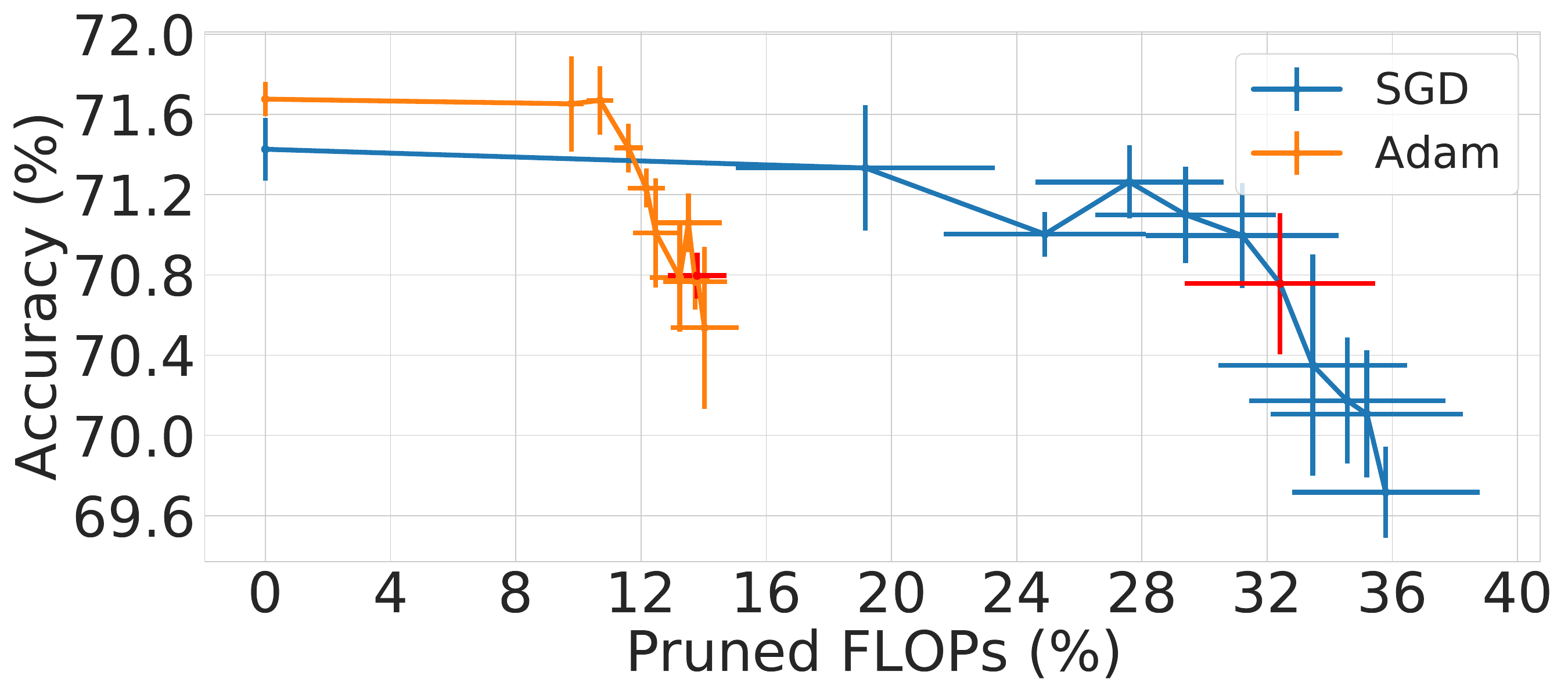}
        \caption{CIFAR-100}
        \label{fig:resnet56_cifar100_flops}
    \end{subfigure}
    \caption{ResNet-56 on CIFAR-10 and CIFAR-100 by iteration averaged over three seeds, the error bars represent standard deviation for these three runs. The red dot corresponds to the selected best iteration.}
    \label{fig:resnet56_cifar}
\end{figure}

For ResNet-20/56 on CIFAR-100 the results are presented in Table \ref{tab:resnet_cifar100_results} and by iteration in Figure \ref{fig:resnet56_cifar}. 

\begin{table}[ht!]
    \begin{center}
        \caption{Results for ResNets trained on CIFAR-100 with SGD and Adam optimizers. During retraining 80 epochs are used; the learning rate is decreased by 10 every 30 epochs. For NNrelief (ours) $\alpha_{\text{conv}} = 0.95, \alpha_{\text{fc}}=0.99$.}
        \label{tab:resnet_cifar100_results}
        \begin{tabular}{  @{} cl p{0.7in}  p{0.7in} p{0.7in} p{0.7in} @{}  } 
        \toprule
        ResNet & Optimizer & Acc $(\%)$ (baseline) & Acc drop $(\%)$ & Parameters pruned (\%) & FLOPs pruned $(\%)$\\ 
        \midrule
        \multirow{2}{0.2cm}{20} 
            & SGD & 68.75 $\pm$ 0.15 & 1.26 $\pm$ 0.33 & 54.54 $\pm$ 0.24 &  12.49 $\pm$ 1.52 \\
            & Adam & 67.34 $\pm$ 0.2 & 0.41 $\pm$ 0.23 & 51.82 $\pm$ 0.23 &  3.42 $\pm$ 1.21 \\
        \midrule
        \multirow{2}{0.2cm}{56} 
            & SGD & 71.43 $\pm$ 0.16 & 0.43 $\pm$ 0.26 & 58.62 $\pm$ 0.26 &  31.2 $\pm$ 3.09 \\
            & Adam & 71.68 $\pm$ 0.09 & 0.88 $\pm$ 0.12  & 67.72 $\pm$ 0.29 &  13.79 $\pm$ 0.94 \\
        \bottomrule
        \end{tabular}
    \end{center}
\end{table}

\begin{figure}[ht]
    \centering
    \includegraphics[width=\textwidth]{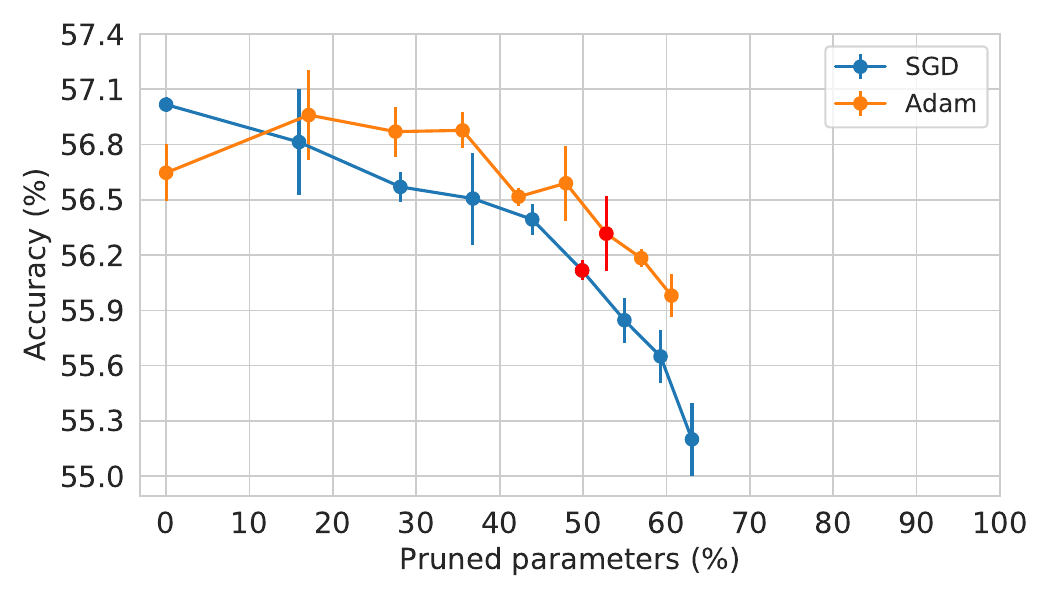}
    \caption{ResNet-18 on Tiny-ImageNet trained with SGD and Adam optimizers. The results are averaged over three random initializations. The bars represent standard deviations considering three runs.}
    \label{fig:resnet18_tiny-imagenet}
\end{figure}

\bmhead{Tiny-ImageNet}Training of a modified ResNet-18 with 16, 32, 64 and 128 output channels indicates that we can prune more than 50\% of the parameters with both optimizers (see Figure \ref{fig:resnet18_tiny-imagenet}).  Adam, however, maintains a higher level of accuracy during retraining than SGD.  We use 2000 samples to evaluate importance scores, and weight decay for (re-)training equal to $5\cdot10^{-4}$.

\bmhead{Sparsity} In contrast to VGG results, we do not observe a significant difference between optimizers in terms of produced sparsity for CIFAR-10/100 and Tiny-ImageNet. In the works that explore the question of sparsity \cite{mehta2019implicit}, ResNet architecture was not explored. It seems that the skip connections allow the signal to pass unhindered to the next layers, making it harder for the Adam optimizer to find which activations are significant, and aggressively prune the less significant connections.

\section{Discussion}

To illustrate the meaning of the importance scores we compare our approach with the magnitude-based one. Also, we present error bounds for the difference of a signal in the layer before and after pruning.

\subsection{Comparison with Magnitude-based approach}

In order to show the difference between our approach and the magnitude-based approach \cite{han2015learning}, we build the heatmaps with the location of top 15.5\% of the most significant connections in the LeNet-300-100 pretrained on MINIST for both our proposed method and the magnitude-based pruning, since this number corresponds to the percentage of parameters that remain after pruning with our proposed rule (importance scores) using $\alpha = 0.95$. The importance scores (IS) are computed from Eq. \ref{eq:fc}, while for the magnitude-based rule, we consider both structured and unstructured pruning. In the structured case, we select the top 15.5\% in each layer independently, and in the unstructured, the top 15.5\% from all layers are selected, meaning that some layers may contain more or less than 15.5\% of the parameters.

\begin{figure}[ht!]
    \begin{subfigure}[b]{\textwidth}
        \centering
        \includegraphics[width=\textwidth]{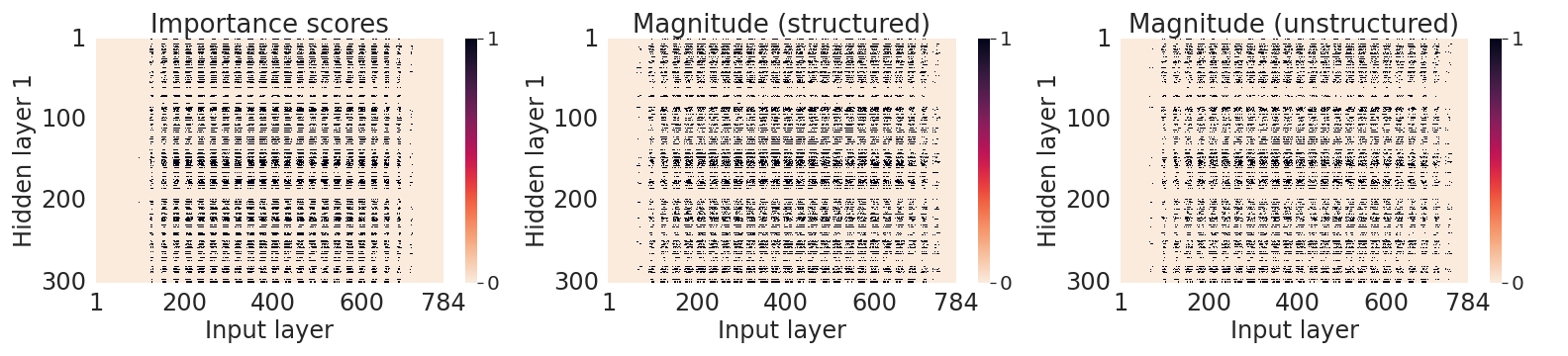}
    \end{subfigure}%
    \vspace{1em}
    \begin{subfigure}[b]{\textwidth}
        \centering
        \includegraphics[width=\textwidth]{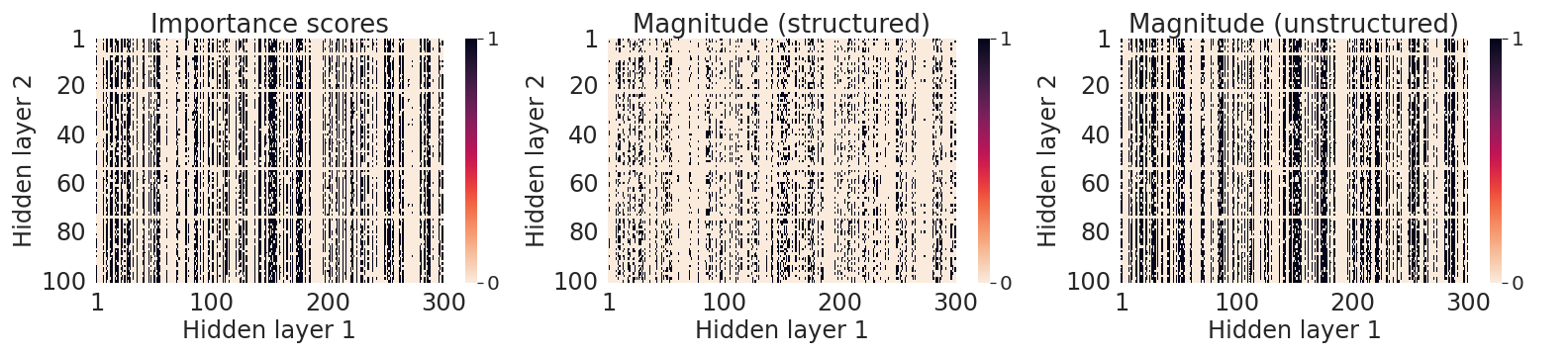}
    \end{subfigure}
    \vspace{1em}
    \begin{subfigure}[b]{\textwidth}
        \centering
        \includegraphics[width=\textwidth]{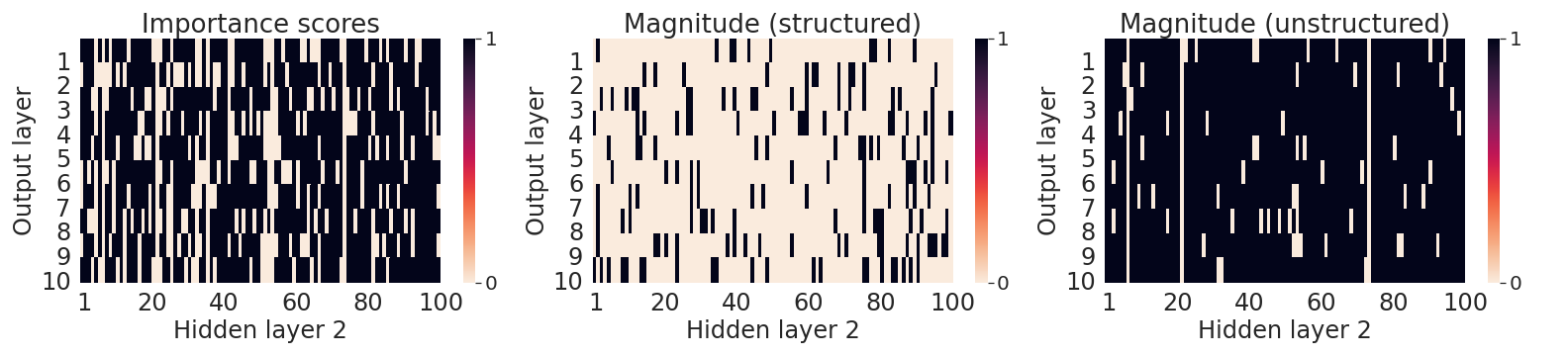}
    \end{subfigure}
    \caption{The remaining 15.5\% connections according to Importance scores and magnitude-based rule (structured and unstructured pruning) for trained LeNet-300-100 on MNIST.}
    \label{fig:comparison_with_magnitude}
\end{figure} 

Figure \ref{fig:comparison_with_magnitude} displays the difference between our approach and the magnitude-based (structured and unstructured) one when determining the significance of network connections. According to the figure, we can see different patterns for the remaining connection, especially for the unstructured magnitude-based rule, where more connections are retained in the last layer compared to the IS rule. Overall, we present the Jaccard index \cite{jaccard1912distribution} between IS and both magnitude-based rules in Table \ref{tab:jaccard_index}. The Jaccard index (or Intersection over Union, IoU) for two sets $A$ and $B$ is defined as follows: $\text{IoU} = \frac{\lvert A \cap B \rvert}{\lvert A \cup B \rvert}$.

\begin{table}[ht]
   \caption{Jaccard index (similarity) between IS rule ($\alpha = 0.95$) and magnitude-based one (structured and unstructured) for each layer if 84.5\% of parameters are pruned in LeNet-300-100 pretrained on MNIST.}
   \centering
    \begin{tabular}{llll}
        \toprule
        & Layer  & Magnitudes (structured) & Magnitudes (unstructured)\\
        \midrule
        \multirow{3}{1.5cm}{Importance scores} 
            & 784 $\to$ 300 & 0.08 & 0.08\\
            & 300 $\to$ 100 & 0.11 & 0.18\\
            & 100 $\to$ 10 & 0.16 & 0.66\\
        \bottomrule    
    \end{tabular}
    \label{tab:jaccard_index}
\end{table}

Both Figure \ref{fig:comparison_with_magnitude} and Table \ref{tab:jaccard_index} demonstrate that the IS rule prunes the network differently from both magnitude-based rules. First, the binary heatmap patterns in the figure indicate a clear difference between the methods. In addition, the IoU indicators are low for every case, meaning that the IS pruning proposed herein determines the most significant connections differently to both magnitude-based rules. From empirical results, we observe that IS-based pruning deactivates more connections in overparameterized layers. In Table \ref{tab:lenet_1iter_arch}, we present the number of active neurons by layer after one pruning iteration (15.5\% of all parameters are retained). As it can be seen, the IS metric deactivates more neurons in every layer compared to unstructured magnitude pruning \cite{han2015learning, frankle2018lottery}. This happens because some neurons produce a low signal (or zero signal because of the ReLU activation function).

\begin{table}[ht]
   \caption{Number of active neurons for magnitude-based pruning and with importance scores one (NNrelief)  for LeNet-300-100 pretrained on MNIST.}
   \centering
    \begin{tabular}{ll}
        \toprule
        Method  & Architecture\\
        \midrule
        Original & $784 \to 300 \to 100 \to 10$ \\
        \midrule
        importance scores & $380 \to 131 \to 96 \to 10$ \\
        magnitude-based (unstructured) & $530 \to 136 \to  97 \to 10$ \\
        \bottomrule    
    \end{tabular}
    \label{tab:lenet_1iter_arch}
\end{table}

In contrast to magnitude-based pruning, NNrelief automatically determines the number of connections to prune when $\alpha$ is given. However, for magnitude-based pruning, the number of pruned connections is a hyperparameter which is difficult to set. For example, if we prune 80\% of the remaining parameters at every iteration as discussed in the Lottery Ticket Hypothesis work \cite{frankle2018lottery}, then after 11 iterations magnitude-based pruning retains $100\% \cdot 0.8^{11} \approx 8.6\%$ of the parameters, while NNrelief retains 1.51\% (see Table \ref{tab:lenets_relu} and Table \ref{tab:pruning}). This is achieved due to the usage of the input signal to estimate the importance scores, which helps to determine the number of connections to prune.

\subsection{Error estimation}\label{sec:error_estimation}

To complement the analysis of our approach we derive error bounds by considering the difference between a signal in the trained neuron and a reduced one without retraining, before and after the activation function. This analysis is important to make sure NNrelief does not destroy the pretrained structure, and that changes in the output neurons are not dramatic enough to prevent successful retraining. To illustrate this, we compare the derived error bounds with the observed changes in ResNet-20 pretrained on CIFAR-10 if the fully connected layer is pruned. Without loss of generality we may assume the first $p$ connections in the neuron $j$ of layer $l$ are kept and that the bias $b^{(l)}_j$ as well as weights $w^{(l)}_{p+1,j}, \ldots, w^{(l)}_{m_lj}$ are pruned in a particular neuron $j$. Then, for any input $x_n^{(l-1)} \in \mathbf{X}^{(l-1)} = \{x_1^{(l-1)}, \ldots, x^{(l-1)}_N\} \subset \mathbb{R}^{m_{l-1}}$ we obtain:

\begin{multline}
\delta_p \big(x_n^{(l-1)}\big) = \Bigg\lvert \sum_{k=1}^{m_{l-1}} w^{(l)}_{kj} x^{(l-1)}_{nk} + b^{(l)}_j - \sum_{k=1}^p w^{(l)}_{kj} x^{(l-1)}_{nk} \Bigg \rvert = \Bigg \lvert \sum_{k=p+1}^{m_{l-1}} w^{(l)}_{kj} x^{(l-1)}_{nk} + b^{(l)}_j \Bigg \rvert \le \\ \le \Big\lvert b^{(l)}_j \Big\rvert + \sum_{k=p+1}^{m_{l-1}} \Big\lvert w^{(l-1)}_{kj} x^{(l-1)}_{nk} \Big\rvert.
\end{multline}

By averaging over all samples:

\begin{multline}
    \label{eq:eq1}
    \overline{\delta_p} \le \Big\lvert b^{(l)}_j \Big\rvert + \sum_{k=p+1}^{m_{l-1}} \overline{\Big\lvert w^{(l)}_{kj}x^{(l-1)}_k} \Big\rvert = S^{(l)}_j \cdot \Bigg(\frac{\Big\lvert b^{(l)}_j \Big\rvert } {S^{(l)}_j} + \frac{\sum_{k=p+1}^{m_{l-1}} \overline{ \Big\lvert w^{(l)}_{kj}x^{(l-1)}_k} \Big\rvert }{S^{(l)}_j}\Bigg) = \\ 
 = S^{(l)}_j \cdot \sum_{k=p+1}^{m_{l-1}} s^{(l)}_{kj} = S^{(l)}_j(1-\alpha).
\end{multline}

Therefore, from \eqref{eq:eq1} we see that the sequence of residuals $\{\overline{\delta_p}\}_{p=1}^m$ decreases with respect to $p$ and increases with $\alpha$. Also, for any Lipschitz continuous activation function $\varphi \in \text{Lip}(X), \ X \subset \mathbb{R}$ with Lipschitz constant $C$ we obtain:

\begin{multline}
    \label{eq:eq2}
    \Delta_p \big(x_n^{(l-1)}\big) = \Big\lvert \varphi \Big(\sum_{k=1}^{m_{l-1}} w^{(l)}_{kj} x^{(l-1)}_{nk} + b^{(l)}_j \Big) - \varphi\Big(\sum_{k=1}^p w^{(l)}_{kj} x^{(l-1)}_{nk}\Big) \Big\rvert \le \\ \le C \Big\lvert \sum_{k=p+1}^{m_{l-1}} w^{(l)}_{kj} x^{(l-1)}_{nk} + b^{(l)}_j \Big\rvert =  C \delta_p \big(x_n^{(l-1)}\big). 
\end{multline}

As a result, from Eq. \eqref{eq:eq1} and Eq. \eqref{eq:eq2}:

\begin{equation}
    \label{eq:fc_ineq}
    \overline{\Delta}_p \leq CS^{(l)}_j(1-\alpha).
\end{equation}

For the point-wise estimation on set $\mathbf{X}^{(l-1)} = \{x_1^{(l-1)}, \ldots, x^{(l-1)}_N\} \subset \mathbb{R}^{m_{l-1}}$, $\forall x_i^{(l-1)}$ we obtain:

\begin{equation}
    \label{eq:fc_ineq_poinwise}
    \Delta_p (x_i^{(l-1)}) \leq \max_{x_n^{(l-1)} \in \mathbf{X}^{(l-1)}} C\delta_p \big(x_n^{(l-1)}\big).
\end{equation}

Since ReLU, ELU, sigmoid and, tanh are Lipschitz continuous functions then \eqref{eq:fc_ineq} and \eqref{eq:fc_ineq_poinwise} hold for them, meaning that the sequence of residuals $\{\overline{\Delta}_p\}_{p=1}^{m_l}$ decreases as well with the increase of $\alpha$. Moreover, in the case of ReLU and ELU, the Lipschitz constant is $C=1$. In summary, by increasing $\alpha$ we indeed reduce the approximation error of the neuron.

Analogously, for input $m_{l-1}$-channeled pruning set $\mathbf{X}^{(l-1)} = \{\mathbf{x}^{(l-1)}_1, \ldots, \mathbf{x}^{(l-1)}_N\}$, where  $\mathbf{x}^{(l-1)}_k = (x^{(l-1)}_{k1}, \ldots, x^{(l-1)}_{km}) \in \mathbb{R}^{m_{l-1} \times h^1_{l-1} \times h^2_{l-1}}$, where $h^1_{l-1}$ and $h^2_{l-1}$ are height and width of input images (or feature maps) for convolutional layer $l-1$, and $\mathbf{B}^{(l)}_j \in \mathbb{R}^{h^1_{l} \times h^2_{l}}$ is a matrix that consists of the same bias value $b^{(l)}_j$ at every element we obtain:

\begin{multline}
    \label{eq:conv_signal_diff1}
    \delta_p \big(x_k^{(l-1)}\big) = \Bigg\lvert\Bigg\lvert \sum_{i=1}^{m_{l-1}}\mathbf{K}^{(l)}_{i}*x^{(l-1)}_{ki} + \mathbf{B}^{(l)}_j - \sum_{i=1}^{p} \mathbf{K}^{(l)}_{i}*x^{(l-1)}_{ki} \Bigg\rvert\Bigg\rvert_F = \\ = \Bigg\lvert\Bigg\lvert \sum_{i=p+1}^{m_{l-1}}\mathbf{K}^{(l)}_{i}*x^{(l-1)}_{ki} + \mathbf{B}^{(l)}_j \Bigg\rvert\Bigg\rvert_F \le 
    \sum_{i=p+1}^{m_{l-1}} \Big\lvert\Big\lvert \mathbf{K}^{(l)}_{i}*x^{(l-1)}_{ki} \Big\rvert\Big\rvert_F + \Big\lvert\Big\lvert \mathbf{B}^{(l)}_j \Big\lvert\Big\lvert_F \le \\ \le \sum_{i=p+1}^{m_{l-1}} \Big\lvert\Big\lvert\mathbf{K}^{(l)}_{i}* \Big\lvert x^{(l-1)}_{ki} \Big\vert \Big\rvert\Big\rvert_F + \sqrt{h^1_l h^2_l} \Big\lvert b^{(l)}_j \Big\rvert.
\end{multline}    

Then, by averaging over all pruning samples similarly to the fully connected case we obtain:

\begin{equation}
    \label{eq:conv_signal_diff2}
    \overline{\delta}_p \le \sum_{i=p+1}^{m_{l-1}} \frac{1}{N}\sum_{k=1}^N \Big\lvert\Big\lvert\mathbf{K}^{(l)}_{i}*x^{(l-1)}_{ki} \Big\rvert\Big\rvert_F = S^{(l)}_j(1-\alpha),
\end{equation}
from which for any Lipschitz continuous activation functions $\varphi$ with Lipschitz constant $C$,  we obtain:

\begin{multline}
    \label{eq:conv_ineq}
    \overline{\Delta}_p = \frac{1}{N}\sum_{k=1}^N \Big\lvert\Big\lvert \varphi\Big(\sum_{i=1}^{m_{l-1}} \mathbf{K}^{(l)}_{i}*x^{(l-1)}_{ki} + \mathbf{B}^{(l)}_j\Big) - \varphi\Big(\sum_{i=1}^{p} \mathbf{K}^{(l)}_{i}*x^{(l-1)}_{ki}\Big) \Big\rvert\Big\lvert_F \le \\ \le CS^{(l)}_j(1-\alpha).
\end{multline}

Inequalities \eqref{eq:fc_ineq}, \eqref{eq:conv_ineq} show how the signal in a neuron or filter, respectively, changes after the layer is pruned. Let us consider how this change influences the network output. We define $N^{(l)}_{i_l}\big(x; \mathbf{W}^{(1:l)}\big)$ as a value in the $i_l^{\text{th}}$ neuron in layer $l$, where $\mathbf{W}^{(1:l)} = (\mathbf{W}^{(1)}, \ldots, \mathbf{W}^{(l)})$ are parameters in layers $1, \ldots, l$. The matrix that we obtain after pruning layer $l$ we define as $\mathbf{\hat{W}}^{(l)}$, and $m_l$ is the number of neurons in layer $l, \ 1 \le l \le L$. Then, for $1 \le l < L$ (for $l=L$ see (\ref{eq:eq1}) in Section 5.2):  

\begin{multline}\label{eq:signal_diff}
    \varepsilon (x) = \Bigg\lvert N^{(L)}_{i_L}\big(x; \mathbf{W}^{(1:L)}\big) - N^{(L)}_{i_L}\big(x; \mathbf{W}^{(1:l-1)}, \mathbf{\hat{W}}^{(l)}, \mathbf{W}^{(l+1:L)}\big) \Bigg\rvert =\\ 
 = \Bigg\lvert\sum_{i_{L-1}=1}^{m_{L-1}}w^{(L)}_{i_{L-1}i_L}\Big(\varphi\big(N^{(L-1)}_{i_{L-1}}\big(x; \mathbf{W}^{(1:L-1)}\big) \big) - \\ - \varphi(N^{(L-1)}_{i_{L-1}}\big(x; \mathbf{W}^{(1:l-1)}, \mathbf{\hat{W}}^{(l)}, \mathbf{W}^{(l+1:L-1)}\big)\big)\Big)\Bigg\rvert \le \\ \le C\sum_{i_{L-1}=1}^{m_{L-1}} \Big\lvert w^{(L)}_{i_{L-1}i_L} \Big\rvert \Big\lvert N^{(L-1)}_{i_{L-1}}\big(x; \mathbf{W}^{(1:L-1)}\big) - \\ - N^{(L-1)}_{i_{L-1}}\big(x; \mathbf{W}^{(1:l-1)}, \mathbf{\hat{W}}^{(l)}, \mathbf{W}^{(l+1:L-1)}\big)\Big\rvert
\end{multline}

Performing similar transformations as in \eqref{eq:signal_diff} until layer $l$, we obtain:

\begin{multline}
    \label{eq:signal_diff2}
    \varepsilon (x) \le  C^{L-l}\sum_{i_{L-1}=1}^{m_{L-1}} \Big\lvert w^{(L)}_{i_{L-1}i_L} \Big\rvert \sum_{i_{L-2}=1}^{m_{L-2}} \Big\lvert w^{(L-1)}_{i_{L-2}i_{L-1}} \Big\rvert \ldots \\ \sum_{i_{l}=1}^{m_{l}} \Big\lvert w^{(l)}_{i_{l}i_{l+1}} \Big\rvert \Big\lvert\varphi\big(N^{(l)}_{i_{i_l}}\big(x; \mathbf{W}^{(1:l)}\big)\big) - \varphi\big(N^{(l)}_{i_{l}}\big(x; \mathbf{W}^{(1:l-1)}, \mathbf{\hat{W}}^{(l)}\big)\big) \Big\rvert
\end{multline}

\begin{figure}[t]
    \centering
    \includegraphics[width=0.9\textwidth]{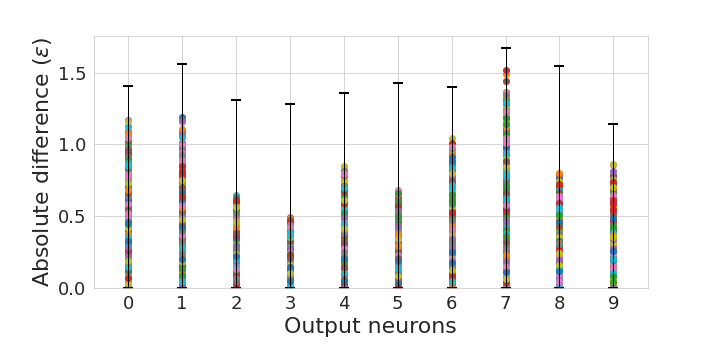}
    \caption{Absolute difference in the neurons' response at the output layer before and after pruning ResNet-20 ($\alpha_{\text{fc}}=0.95$). The bars represent the theoretical bounds, and dots present empirical observations for 10000 images.}
    \label{fig:resnet20_error_estim}
\end{figure} 

Then, averaging \eqref{eq:signal_diff2} over all pruning samples from $\mathbf{X}$ and using \eqref{eq:fc_ineq}:

\begin{equation}
\label{eq:signal_diff2_avg}
\begin{split}
    \overline{\varepsilon} \le C^{L-l}\sum_{i_{L-1}=1}^{m_{L-1}}  \Big\lvert w^{(L)}_{i_{L-1}i_L} \Big\rvert \sum_{i_{L-2}=1}^{m_{L-2}} \Big\lvert w^{(L-1)}_{i_{L-2}i_{L-1}}\Big\rvert \ldots \sum_{i_{l}=1}^{m_{l}} \Big\lvert w^{(l)}_{i_{l}i_{l+1}}\Big\rvert \overline{\Delta}^{(l)}_{p_{i_{l}}} \le \\ (1-\alpha) C^{L-l}\sum_{i_{L-1}=1}^{m_{L-1}} \Big\lvert w^{(L)}_{i_{L-1}i_L} \Big\rvert \sum_{i_{L-2}=1}^{m_{L-2}} \Big\lvert w^{(L-1)}_{i_{L-2}i_{L-1}} \Big\rvert \ldots \sum_{i_{l}=1}^{m_{l+1}} \Big\lvert w^{(l)}_{i_{l}i_{l+1}}\Big\lvert S^{(l)}_{i_l}.
\end{split}    
\end{equation}

Inequality \eqref{eq:signal_diff2_avg} shows the bounds of the network error change after pruning on the pruning set $\mathbf{X}$. Figure \ref{fig:resnet20_error_estim} shows the absolute changes in output responses for ResNet-20 after pruning on CIFAR-10 with $\alpha_{\text{fc}} = 0.95$. The error bars (in black) show the theoretical bounds of the difference estimated in \eqref{eq:fc_ineq_poinwise} and \eqref{eq:signal_diff2}, and the dots demonstrate the actual changes computed on the test data. From the figure, one can observe that the theoretical bounds capture the signal changes well in most cases, especially, for output neurons 0, 6 and 7.

\section{Conclusion}

We propose a new pruning strategy and associated error estimation that is applicable to different neural network architectures. Our algorithm is based on eliminating layers' kernels or connections that do not contribute to the signal in the next layer. In this method, we estimate the average signal strength coming through each connection via proposed importance scores and try to keep the signal level in the neurons close to the original level. Thus, the number of removed parameters in a layer is selected adaptively, rather than being predetermined. As a result, the algorithm obtains a subnetwork capable of propagating most of the original signal while using fewer parameters than other reported strategies. We explore the effect of Adam and SGD optimizers on our pruning strategy. The results show that for VGG architecture Adam optimizer gives much higher compression than SGD, however, for ResNet, we do not observe similar behaviour. We observe that the obtained subnetwork tends to homogenize connection importance, hinting that every remaining connection is approaching similar importance for the dataset on average.

\backmatter

\begin{appendices}

\section{Pruning setups}
\label{sec:prunig_setups}

We implement our approach with PyTorch \cite{paszke2019pytorch}. Table \ref{tab:experiments} shows the training parameters for each network.

\begin{table}[h!]
\centering
 \caption{Training setups.}
 \label{tab:experiments}
 \begin{tabular}{ @{} p{0.8in} p{0.75in} l l p{0.35in} @{}}
 \toprule
 Network & dataset & optimizer & learning rate (by epoch) & weight decay \\
 \midrule
 LeNet-5/300-100 & MNIST & Adam &
{\tiny
$
\begin{cases}
10^{-3}, & 1 \le epoch \le 30\\
10^{-4}, & 31 \le epoch \le 60
\end{cases}$} & $5 \cdot 10^{-4}$\\[0.4cm] 
 VGG-like/13 & CIFAR-10/100, Tiny-ImageNet & SGD/Adam & {\tiny
                                    $
                                    \begin{cases}
                                    10^{-1}/10^{-3}, & 1 \le epoch \le 80\\
                                    10^{-2}/10^{-4}, & 81 \le epoch \le 120\\
                                    10^{-3}/10^{-5}, & 121 \le epoch \le 150
                                    \end{cases}$} & $5 \cdot 10^{-4}$\\[0.4cm]
 \multirow{4}{2cm}{ResNet-20/56} & \multirow{4}{2cm}{CIFAR-10/100} 
                                & SGD &
                                {\tiny
                                    $
                                    \begin{cases}
                                    0.1, & 1 \le epoch \le 80\\
                                    0.01, & 81 \le epoch \le 120\\
                                    0.001, & 121 \le epoch \le 150
                                    \end{cases}$}& $5 \cdot 10^{-4}$\\[0.4cm]
                                && Adam & 
                                {\tiny
                                    $
                                    \begin{cases}
                                    10^{-3}, & 1 \le epoch \le 120\\
                                    10^{-4}, & 121 \le epoch \le 160\\
                                    10^{-5}, & 161 \le epoch \le 200
                                    \end{cases}$}
                                    & $5 \cdot 10^{-4}$\\[1ex] 
 \botrule
 \end{tabular}
\end{table}

Table \ref{tab:retraining} shows the parameters for retraining on every iteration.

\begin{table}[ht!]
\centering
 \caption{Retraining parameters}
 \label{tab:retraining}
 \begin{tabular}{ @{} p{0.85in} p{0.75in} l l p{0.35in} @{}} 
 \toprule
 Network & dataset & optimizer & learning rate (by epoch) & weight decay \\
 \midrule
 LeNet-5/300-100 & MNIST & Adam & {\tiny
$
\begin{cases}
10^{-3}, & 1 \le epoch \le 30\\
10^{-4}, & 31 \le epoch \le 60
\end{cases}$} & $5 \cdot 10^{-4}$\\[0.4cm]
VGG-like/13 & CIFAR-10/100, Tiny-ImageNet & SGD/Adam &{\tiny
                                    $
                                    \begin{cases}
                                    10^{-1}/10^{-3}, & 1 \le epoch \le 20\\
                                    10^{-2}/10^{-4}, & 21 \le epoch \le 40\\
                                    10^{-3}/10^{-5}, & 41 \le epoch \le 60
                                    \end{cases}$}& $5 \cdot 10^{-4}$\\[0.4cm]
 \multirow{4}{2cm}{ResNet-20/56}  
& CIFAR-10 & SGD/Adam &{\tiny
                        $
                        \begin{cases}
                            10^{-1}/10^{-3}, & 1 \le epoch \le 20\\
                            10^{-2}/10^{-4}, & 21 \le epoch \le 40\\
                            10^{-3}/10^{-5}, & 41 \le epoch \le 60
                        \end{cases}$}& $5 \cdot 10^{-4}$\\[0.4cm]
& CIFAR-100 & SGD/Adam &{\tiny
                        $
                        \begin{cases}
                            10^{-1}/10^{-3}, & 1 \le epoch \le 30\\
                            10^{-2}/10^{-4}, & 21 \le epoch \le 60\\
                            10^{-3}/10^{-5}, & 41 \le epoch \le 80
                        \end{cases}$}& $5 \cdot 10^{-4}$\\[1ex] 
 \botrule
 \end{tabular}
\end{table}

Table \ref{tab:pruning} shows the pruning parameters and total number of iterations.

\begin{table}[ht!]
 \caption{Pruning parameters}
 \label{tab:pruning}
\centering
 \begin{tabular}{@{} llll p{0.7in} @{}} 
 \toprule
 Network & dataset & $(\alpha_{\text{conv}}$, $\alpha_{\text{fc}})$ & \# iterations & best iteration SGD/Adam \\ [0.5ex]
 \midrule
 LeNet-300-100 & MNIST & (-, 0.95) & 15 & 11\\ 
 LeNet-5 & MNIST & (0.9,0.95) & 20 & 20\\ 
 \multirow{2}{1.3cm}{VGG-like} 
 & CIFAR-10 & (0.95,0.95) & 6 & 6\\
 & Tiny-ImageNet & (0.95,0.95) & 5& 5\\
 \multirow{2}{3.5cm}{ResNet-20 (SGD/Adam)}  
& CIFAR-10 & (0.95,0.99) & 10 & 7/10\\
& CIFAR-100 &  (0.95,0.99) & 10 & 7/8\\
\multirow{2}{3.5cm}{ResNet-56 (SGD/Adam)}  
& CIFAR-10 & (0.95,0.99) & 10 & 6/2\\
& CIFAR-100 &  (0.95,0.99) & 10 & 6/9\\[1ex] 
 \botrule
 \end{tabular}
\end{table}

\section{FLOPs computation}
\label{sec:flops}
According to \cite{molchanov2016pruning}, we compute FLOPs as follows:
\begin{itemize}
    \item for a convolutional layer: FLOPs $= 2HW(C_{in}K^2+1)C_{out}$ where $H, W$ and $C_{in}$ are height, width and number of channels of the input feature map, $K$ is the kernel width (and height due to symmetry), and $C_{out}$ is the number of output channels.
    \item for a fully connected layer: FLOPs $= (2I-1)O$, where $I$ is the input dimensionality and O is the output dimensionality.
\end{itemize}




\end{appendices}



\bibliography{sn-bibliography}


\end{document}